\documentclass[journal,transmag,twoside]{IEEEtran}

\usepackage{times}
\usepackage{multicol}
\usepackage[hidelinks]{hyperref} 

\usepackage[numbers]{natbib}
\bibpunct{[}{]}{,}{n}{}{;}
\usepackage[noadjust]{cite}  

\usepackage{graphicx}
\usepackage{setspace}
\usepackage{array}
\usepackage{longtable}
\usepackage{multirow}
\usepackage{hhline}
\usepackage{rotating}
\usepackage{amssymb, amsmath, mathrsfs}
\usepackage{color}
\graphicspath{{./images/}}
\DeclareGraphicsExtensions{.pdf,.png,.jpg}
\usepackage{algorithm}
\usepackage{algorithmic}
\usepackage{subfigure}
\usepackage{tabularx}
\usepackage{multirow}

\newcounter{tecounter}
\definecolor{orange}{rgb}{1,0.2,0}
\newcommand{\jc}[1]{{{#1}}} 

\begin{document}	
\title{Towards Automatic Manipulation of Intra-cardiac Echocardiography Catheter}

\author{\IEEEauthorblockN{Young-Ho Kim\IEEEauthorrefmark{1}\IEEEauthorrefmark{*},
	Jarrod Collins\IEEEauthorrefmark{1}\IEEEauthorrefmark{*},
	Zhongyu Li\IEEEauthorrefmark{2},
	Ponraj Chinnadurai\IEEEauthorrefmark{3}, 
	Ankur Kapoor\IEEEauthorrefmark{1},  \\
	C. Huie Lin\IEEEauthorrefmark{2}\IEEEauthorrefmark{,4}, and
	Tommaso Mansi\IEEEauthorrefmark{1}}
	
	\IEEEauthorblockA{\IEEEauthorrefmark{1}Siemens Healthineers, Digital Technology \& Innovation, Princeton, NJ, USA}
	\IEEEauthorblockA{\IEEEauthorrefmark{2}Houston Methodist Research Institute, Houston, TX, USA}
	\IEEEauthorblockA{\IEEEauthorrefmark{3}Siemens Medical Solutions Inc., Advanced Therapies, Malvern, PA, USA}
	\IEEEauthorblockA{\IEEEauthorrefmark{4}Houston Methodist DeBakey Heart \& Vascular Center, Houston, TX, USA}
	\thanks{\IEEEauthorrefmark{*}Young-Ho Kim and Jarrod Collins are co-first authors. Corresponding author: Young-Ho Kim (young-ho.kim@siemens-healthineers.com).}
}

\IEEEtitleabstractindextext
{
\begin{abstract}
\jc{Intra-cardiac Echocardiography (ICE) is a powerful imaging modality for guiding electrophysiology and structural heart interventions. ICE provides real-time observation of anatomy, catheters, and emergent complications. However, this increased reliance on intraprocedural imaging creates a high cognitive demand on physicians who can often serve as interventionalist and imager. We present a robotic manipulator for ICE catheters to assist physicians with imaging and serve as a platform for developing processes for procedural automation. Herein, we introduce two application modules towards these goals: (1) a view recovery process that allows physicians to save views during intervention and automatically return with the push of a button and (2) a data-driven approach to compensate kinematic model errors that result from non-linear behaviors in catheter bending, providing more precise control of the catheter tip. View recovery is validated by repeated catheter positioning in cardiac phantom and animal experiments with position- and image-based analysis. 
We present a simplified calibration approach for error compensation and verify with complex rotation of the catheter in benchtop and phantom experiments under varying realistic curvature conditions. Results support that a robotic manipulator for ICE can provide an efficient and reproducible tool, potentially reducing execution time and promoting greater utilization of ICE imaging. }

\vspace{-10pt}
\end{abstract}
\begin{IEEEkeywords}
	Intra-cardiac echocardiography (ICE), Catheter, Continuum manipulator, Tendon-driven manipulator, Path planning, Automated View Recovery, Non-linear elasticity compensation, Cardiac Imaging \vspace{-15pt}
\end{IEEEkeywords}
}

\maketitle

\section{Introduction}


Interventional cardiology has expanded its role dramatically in recent years to now encompass treatment of many disease states which were once considered to only have surgical options. This growth has been significantly motivated by the introduction of new treatment devices and advances in intraoperative imaging modalities. Intra-cardiac echocardiography (ICE) has been evolving as a real-time imaging modality for guiding interventional procedures in electrophysiology\,\citep{epstein1998ep, daoud1999ep, calo2002ep, basman2017shd}, congenital\,\citep{rigatelli2005chd, tan2019chd}, and structural heart interventions\,\citep{basman2017shd}, among others. 
When compared to another more established real-time imaging modality, transesophageal echocardiography (TEE), ICE has improved patient tolerance by not requiring esophageal intubation, requires only local anesthesia with conscious sedation, \jc{can be operated by the interventionalist}, and does not interfere with fluoroscopic imaging\,\citep{silvestry2009ice}. 
Real-time ICE imaging has an expanding role in providing uninterrupted guidance for valve replacement interventions\,\citep{green2004mv, bartel2011tavi, ahmari2012mv, marmagkiolis2013mv,  henning2014mv, bartel2016tavi, saji2016mv, patzelt2017mv}, left atrial appendage closure\,\citep{rao2005laa, shah2008laa, ren2013laa, anter2014laa, berti2014laa, matsuo2016laa}, septal defect closure\,\citep{hijazi2001asd, mullen2003asd, medford2014asd}, and catheter-based ablation for cardiac arrhythmia\,\citep{kalman1997abl, saliba2008abl, bhatia2010abl, filgueiras2015abl}. However, with the increased reliance on imaging to perform these complex procedures, there is a high cognitive demand on physicians, who \jc{may now be} performing both the interventional task and simultaneously acquiring the guiding images. Moreover, they are not \jc{always} experts in reading ultrasound and navigating these images, which makes ICE handling even more difficult.

ICE imaging requires substantial training and experience to become comfortable with steering the catheter \jc{from within the cardiac anatomy}, which hinders its adoption as standard of care\,\citep{vitulano2015ice, bartel2016tavi, tan2019chd}. In practice, the interventionalist is needing to continuously manipulate several catheters throughout the procedure, each having different control mechanisms. 
For example, a typical ablation treatment for cardiac arrhythmia can require tens to hundreds of individual ablations at very specific locations. ICE imaging can be beneficial \jc{to monitor} for developing complications, target anatomy, facilitate adequate tissue contact, and monitor lesion development during ablations\,\citep{saliba2008abl}. However, this can become an iterative and time-consuming procedure when frequently needing to re-position both the therapeutic and imaging catheters. \jc{Similarly} in structural heart procedures, clinicians can manipulate the ICE catheter to localize and measure the area of treatment and then either \textit{park} (\textit{e.g.} to watch for complications) or \textit{retract} the ICE imaging catheter while devices are deployed under fluoroscopic guidance. The ICE catheter is then relocated to visually confirm the placement of therapeutic devices. This manner of repeated manipulation throughout the course of treatment is common for interventions across disciplines but requires intensive coordination, spatial understanding, and manual dexterity that can lead to fatigue in longer or more difficult procedures and imposes a significant learning curve for new users. 

When considering these limitations, it is \jc{reasonable} that a robotic-assist system to hold and actively manipulate the ICE catheter, either through operator input or \jc{semi-autonomous processes}, could ease the workload of the physician during treatment and potentially enable the use of ICE for novel and more complex tasks. 
Several commercial robotic systems for \jc{less specific} catheter manipulation are currently marketed, including Amigo RCS (Catheter Precision, Inc., Mount Olive, NJ, USA), CorPath GRX (Corindus Inc., Waltham, MA), Magellan, and Sensei (Hansen Medical Inc., Mountain View, CA, USA).
One commercially-available robotic system for ICE catheter manipulation is the Sterotaxis V-Sono system\,\citep{stereotaxis20}, which controls the ICE catheter robotically, but with reduced degrees-of-freedom. \jc{This system provides} robotic control of devices \jc{via} human operators at a remote cockpit based on \jc{streamed} real-time image (\textit{e.g.} fluoroscopic) feedback. 
\citet{loschak16ice} have presented a \jc{research prototype} robotic ICE manipulator and provide a method using electromagnetic (EM) tracking systems to actively maintain focus within the field of view. In further work\,\citep{loschak17predictive} they apply a filtering method from EM sensors to compensate for respiratory motion.
However, traditional \jc{position-tracking sensors}, like EM, cannot always be mounted at the tip of the catheter to provide required feedback due to practical limitations ({\em e.g.} cost, size, sterilization).
While some ICE catheters are manufactured with an EM sensor in the tip for application in EP procedures, there is \jc{an} accompanying increase in cost.
In practice, many commercially available ICE catheters are single use with no \jc{position-tracking} sensors installed. Therefore, the controller for such a robotic-assist system requires an open-loop where spatial feedback are not continuously available. \jc{Accordingly, with this work we introduce a robotic ICE catheter controller that operates without discrete position feedback from added sensors, demonstrate a control application for automating ICE imaging during procedures, and apply methods for compensating non-linear behaviors in catheter control to improve precision.}

\begin{figure}[t!]
	\centering
	\includegraphics[scale= 0.45]{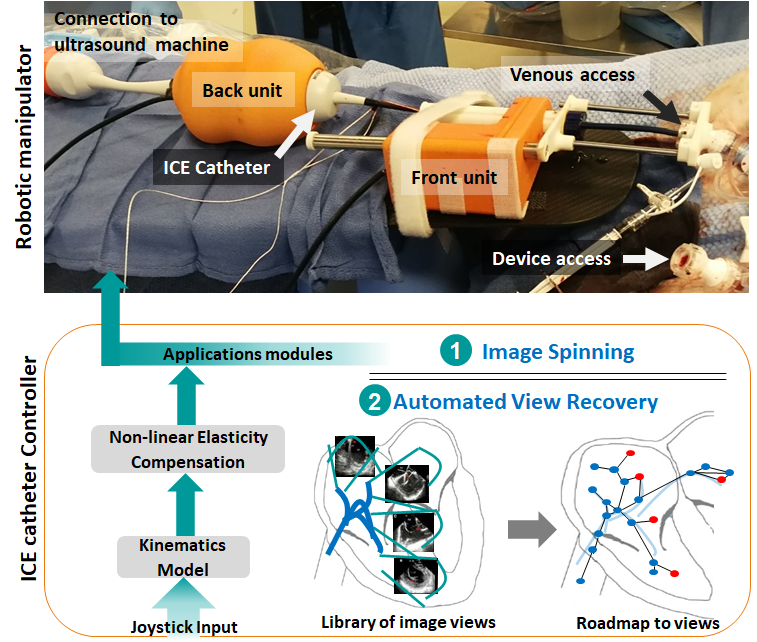}
	\vspace*{-5pt}
	\caption{\jc{An overview of the proposed robotic ICE manipulator and control scheme. Joystick, or similar, input can direct the robotic manipulator. The non-linear elasticity compensation models presented in this paper are applied to joystick input to provide more smooth and precise control. Alternatively, application modules (\textit{e.g.} image spinning and automated view recovery) can provide intelligent instructions for the robotic manipulator to achieve a task. In this work, the non-linear elasticity compensation is applied to the image spinning task but not the automated view recovery.}\label{fig:overview}}
	\vspace*{-15pt}
\end{figure}


\jc{Herein, we present new methods to simplify ICE catheter manipulation for the interventionalist that are enabled by our robotic ICE catheter controller. Figure\, \ref{fig:overview} shows an overview of the robotic-assist system. More specifically, and enabled by the robotic controller, we present the following:} 
\begin{enumerate}
    \item \jc{As a first step towards automation in clinical ICE workflows, we propose an ultrasound view recovery method which can autonomously reproduce important views previously saved by the user. To achieve, we implement methods to incrementally generate a topological map of motor control states. When queried, the method provides a path to a specified view that can be reproduced at any time by the robotic controller.}
    \item \jc{In working towards more precise robotic control, which will be fundamental when developing further autonomous control applications, we apply data-driven models to compensate catheter kinematics for highly non-linear elasticity behaviors caused by the structural composition of the ICE catheter; Specifically, an interpolation-based mapping function is applied with a clinically-practical calibration method.}
\end{enumerate}
\jc{We present evaluation of these proposed methods in a combination of benchtop, phantom, and animal experiments.}

\section{Background and Related works}\label{sec:relatedwork}
\subsection{Kinematics and non-linear elasticity models of steerable catheters}

An ICE catheter is a long, thin, and flexible structure, categorized as a continuum device. Two pairs of tendon mechanisms in the backbone control the ICE catheter tip, \jc{creating} an underactuated system. Use of this flexible structure in practical applications requires models of robotic shape and motion, which is more complex than traditional rigid body robots.
Several approaches exist to characterize positions, forces, and moments of the tip by actuators and external interactions with the environment. 
However, due to the complexities of modeling, most methods are solved numerically. As such, many publications have presented a simplified approach. The piecewise constant curvature approximation models the robot as a series of mutually tangent constant-curvature arcs\,\citep{webster10ccc}, therefore providing closed-form kinematics and Jacobian formulation. There exist two mapping stages: one is from joint space to configuration parameters that describe constant-curvature arcs and the other mapping is from these configuration parameters to task space, consisting of a space curve which describes position and orientation along the backbone (detailed in Figure~2-3 in \,\citet{webster10ccc}).
Many continuum manipulators related to medical applications ({\em
e.g.} endoscopes\,\citep{ott11endoscopy,kao14tendon,do14hysteresis,xu17tendon}, colonoscopes\,\citep{chen06kinematics}, catheters\,\citep{loschak16ice}) are composed of arcs. While imperfect, the constant curvature models are widely applied due to remarkable usefulness of the model approximation. 
However, the existing ICE catheter is controlled by multiple tendons, which are coupled and have non-linear tensions due to polymer effects (detailed in Section\,\ref{sec:icecatheter}). 
\jc{As a result, the assumptions of the constant curvature model are invalidated; the effects of non-linear elasticity are neglected. Therefore, additional mapping functions are required to have concise control of the tip. 
There exist works\,\citep{kai08continuum,simaan10design} related to non-linearity compensation of the constant curvature model in bending plane. The fitting function in configuration space compensates the angle errors of bending plane and the errors of out of plane due to non-linearity behavior.
Similarly, we construct a kinematic mapping function; however based on interpolation-based approach with a simplified calibration method.
}



\subsection{Ultrasound Imaging and Automation}

For some time, physicians have applied ultrasound imaging to observe, detect, and track target anatomies or surgical tools\,\citep{priester13us,antico19us} to great effect in a wide variety of procedures. Ultrasound image-based (semi-) autonomous robotic systems have been studied extensively for use in tracking the prostate for brachytherapy\,\citep{long12prostate}, to aid in target visualization and tool positioing for liver ablation\,\citep{boctor08us,xu10us}, to detect the tumour boundary in partial nephrectomy\,\citep{papalia12nephrectomy}, and for catheter tracking in multimodal imaging\,\citep{wu15catheter}.

Robotic catheter systems have been developed to improve maneuverability\,\citep{loschak16ice}, compensate for the heart motion\,\citep{loschak17predictive}, and mapping for catheter navigation\,\citep{issa19mapping}. However, there are still many interventional processes that can benefit from the fast, repeatable, and precise controls that robotic manipulation can provide. Accordingly, we expect that the automated view recovery method presented in this work to be of high value to clinical ICE imaging.




\section{Materials and Methods}\label{sec:tech}
First, we present the ICE catheter mechanism, our robotic system, and we re-visit the existing kinematics models with our proposed compensation method. Then, we introduce our separate view recovery method.

\subsection{ICE catheter mechanism}\label{sec:icecatheter}

\begin{figure}[t!]	
	\begin{center}\hspace*{-0pt}
	\includegraphics[scale= 0.35]{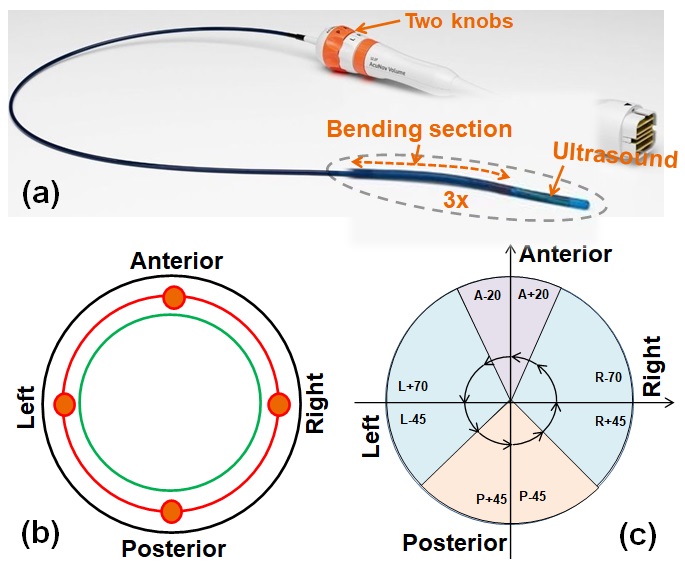}
	\end{center}\vspace*{-0pt}
	\caption{(a) The ICE catheter is shown with two knobs, bending section, and ultrasound array labeled. (b) Catheter cross-section: four thread sections (orange) combining with polymer cover (black) and ultrasound array section (green), (c) Mechanical tolerance requirement   \label{fig:catheter_overall}
	}
	\vspace*{-0pt}
\end{figure}

The ICE catheter structure has a highly nonlinear behavior due to various slack (dead-zone), elasticity, and hysteresis phenomena in multiple coupled components involved in tendon control. 
The mechanical composition of the ICE catheter (Figure\,\ref{fig:catheter_overall}~(a)) is based on two pairs of tendon-sheath pull mechanisms, which consist of a hollow polymer as a sheath and a thread sliding inside the sheath acting as a tendon. Each pair is bound to a common knob, which can pull an individual thread by rotating the knob, allowing one thread to be pulled while the other remains passive. This structure assumes ideally zero-slack transition between threads; however, this is not realistically achievable.
Moreover, the ultrasound array is located in the center of the two pairs of tendon mechanisms which are covered by the outer polymer shell for sterilization purposes (Figure\,\ref{fig:catheter_overall}~(b)). 
Highly nonlinear behaviors exist due to these structural considerations. \jc{While there are multiple contributing phenomena, we focus on non-linear elasticity compensation in bending structures; the configuration parameters of constant curvature models do not represent the real tip pose due to uncertainty in the bending section. This leads to the extreme mechanical tolerances presented in Figure\,\ref{fig:catheter_overall}~(c).}


Precise prediction of the catheter tip pose for a specific knob configuration is therefore challenged by these non-linear properties. EM tracking systems have been used to co-localize the tip of the catheter in space. However, this increases the cost of the catheter and is currently only available, or clinically used, for EP procedures. In this paper, we investigate an open-loop control to address these challenges.

\subsection{ICE catheter kinematics}\label{sec:robot design}

We developed an ICE catheter robotic control system for this study.  
Our robot manipulator consists of two components as shown in Figure\,\,\ref{fig:overview}:
1) A {\it ``Front"} component holds the catheter shaft, sits directly outside of the introducer sheath, and contributes linear and rotational motion of the catheter. 2) A {\it ``Back"} component holds the catheter handle and controls the two knobs for the bending of the catheter tip, and bulk rotation of the catheter. Moreover, this bulk rotation is synchronized with the \textit{Front}.

The robot has 4 degrees of freedom. Without loss of generality, we follow the same nomenclature from\,\citep{loschak16ice}: two DOFs for steering the catheter tip in two planes (anterior-posterior knob angle $\phi_1$ and right-left knob angle $\phi_2$) using two knobs on the catheter handle, bulk rotation $\phi_3$, and translation $d_4$ along the major axis of the catheter. We define the robot's configuration, ${\bf q} = (\phi_1,\phi_2,\phi_3,d_4)$ in $\mathbb{R}^4$. 
The robot can be controlled by manually using an external joystick, which provides a digital input that is directly mapped to the standard knob controls of the catheter or a more intuitive control scheme where the users inputs are directly applied at the catheter tip coordinate frame. 


\begin{figure}[t!]	
	\begin{center}\hspace*{-0pt}
     \includegraphics[scale= 0.75]{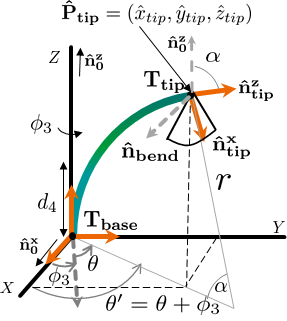}\label{fig:fk}
	\end{center}\vspace*{-10pt}
	\caption{Catheter bending geometry and kinematics nomenclature
		\label{fig:kinematics}}
	\vspace*{-10pt}
\end{figure}

For completeness, we summarize the explicit closed-form kinematics model based on \,\citep{loschak16ice}. 
The overall catheter bending geometry is described in Figure\,\ref{fig:kinematics}. To use constant curvature models, we setup the base coordinate frame $T_{base}$ for the bottom of the bending section, and $T_{tip}$ is the catheter tip coordinate frame where the ultrasound array is installed and the center of the image facing direction is ${\bf \hat{n}}_{tip}^x$. The bending section length is defined as $L$. 

\vspace*{8pt}
{\noindent \it Forward Kinemaetics (FK) from ${\bf q}$ to $T_{tip}$:}\\
There exist two configuration parameters in the constant curvature model, $\theta$ and $\alpha$; $\theta$ is the right-handed rotation angle from ${\bf \hat{n}}_0^x$ (the $x$-axis of $T_{base}$) along ${\bf\hat{n}}_0^z$ (the $z$-axis of $T_{base}$); $\alpha$ is the the angle of the curvature, which is computed from anterior-posterior thread deflection, $L_{ap}$, and right-left thread deflection, $L_{rl}$. These thread deflections are normally computed from $\phi_1$ and $\phi_2$ with a knob radius $r_{knob}$ as $L_{ap} = \phi_1 \cdot r_{knob}$ and $L_{rl} = \phi_2 \cdot r_{knob}$. Then, $\alpha$ can be computed as $\sqrt{(\frac{L_{ap}}{r_{catheter}})^2 + (\frac{L_{rl}}{r_{catheter}})^2}$, where $r_{catheter}$ is a catheter radius. 





The remainder parts of kinematics as follows:
\begin{eqnarray}
\theta = tan^{-1}(\frac{\phi_2}{\phi_1}),~r =\frac{L}{\alpha}\label{eg:fkTheta}\\
\begin{aligned}
\hat{x}_{tip}=r~(1-cos~\alpha)cos~\theta,\\
\hat{y}_{tip}=r~(1-cos~\alpha)sin~\theta,\\
\hat{z}_{tip}=r~sin~\alpha,\label{eg:fkPosition}
\end{aligned}
\end{eqnarray}

More specifically, $\theta$ is the angle between the bending plane (due to $\phi_1$ and $\phi_2$) and the X-Z plane when $\phi_3=0$ (Figure\,\ref{fig:kinematics}). $\phi_3$ will be added later. $r$ is the radius of curvature in Equation\,\eqref{eg:fkTheta}. Then, the catheter tip position ${\bf \hat{P}_{tip}}$ is calculated from Equation\,\eqref{eg:fkPosition}.

The orientation of the tip can be calculated by the {\it Rodrigues' rotation formula}, which is a method for rotating a vector in space, given an axis and angle of rotation. Let $R({\bf \mu}, \beta)$ be the rotation matrix using {\it Rodrigues' rotation formula}  from the given axis ${\bf \mu}$ and rotating it by an angle $\beta$ according to the right hand rule. 
Then, the orientation of the tip is computed by $R({\bf \hat{n}}_{bend}, \alpha)$, where {${\bf \hat{n}}_{bend}$} is the vector orthogonal to the bending plane.



Let ${T_{tilt}}$ be the 4x4 transformation matrix from ${\bf \hat{P}_{tip}}$ and  $R({\bf \hat{n}}_{bend}, \alpha)$  without the body rotation $\phi_3$ and translation $d_4$. The rotation of $\phi_3$ is $T_{roll}(\phi3)$. The translation of $d_4$ is $T_{trans}(d_4)$. Then, the overall transformation matrix $T_{tip}$ is given as $T_{trans}(d_4)\cdot T_{roll}(\phi3) \cdot T_{tilt} \cdot T_{US}$ where $T_{US}$ is the constant transformation from the tip of the bending section to the center of the ultrasound images along the catheter axis.

\vspace*{8pt}
{\noindent \it Inverse Kinematics (IK) from ${\bf T_{tip}}$ to ${\bf q}$:}\\
Similarly, a closed-form solution of the inverse kinematics is as follows based on \,\citep{loschak16ice}:
\begin{eqnarray}
\alpha = cos^{-1}({\bf \hat{n}}_0^z \cdot {\bf \hat{n}}_{tip}^{z})\label{eq:ik_alpha},\\
\theta' = atan2(\hat{y}_{tip}, \hat{x}_{tip}),\label{eq:ik_thetadot}\\
\hat{\phi}_3 = \theta' - atan2({\bf\hat{n}}_{tip}^{x} \times  R({\bf \hat{n}}_0^{x},\alpha), {\bf \hat{n}}_{tip}^{x}\cdot R({\bf \hat{n}}_0^{x},\alpha) ), \\
\hat{\phi}_1 = \frac{\alpha \cdot r_{catheter} \cdot cos (\theta)}{r_{knob}},\\
\hat{\phi}_2 = \frac{\alpha \cdot r_{catheter} \cdot sin (\theta)}{r_{knob}},\\
\hat{d_4} = \hat{z}_{tip} - r\cdot sin\alpha,
\end{eqnarray}

$\alpha$ is the dot product of ${\bf\hat{n}}_0^z$ (the $z$-axis of $T_{base}$) and ${\bf\hat{n}}_{tip}^{z}$ (the $z$-axis of catheter tip) in Equation\,\eqref{eq:ik_alpha}. The $\theta'$ is computed from the catheter tip position ${\bf \hat{P}_{tip}}$ in Equation\,\eqref{eq:ik_thetadot}. $\hat{\phi}_3$ is computed from the angle between $R({\bf\hat{n}}_0^{x},\alpha)$ ($x$-axis of $T_{base}$ rotated by $\alpha$) and the dot product of ${\bf\hat{n}}_{tip}^{x}$ ($x$-axis of $T_{tip}$) and $R({\bf\hat{n}}_0^{x},\alpha)$. 
The computed $\hat{\phi}_1$ and $\hat{\phi}_2$ represent the estimated joint states.


 \subsection{Non-linear elasticity compensation}\label{sec:compensator}
 
\jc{The original model assumes that the bending section angles ($\alpha$ and $\theta$) are purely determined by linear thread deflections $L_{ap}$ and $L_{rl}$. However, two types of error result from non-linear elasticity in the bending structures: 1) The real bending plane is not located in the computed $\theta$ and 2) the angle of bending plane $\alpha$ is non-linear with regard to $L_{ap}$ and $L_{rl}$, which varies with each unique catheter. Thus, we need to compensate $\hat{\phi}_1$ and $\hat{\phi}_2$ for the updated real values $\phi'_1$ and $\phi'_2$.}

\jc{We note that other estimated joint states ($\phi_3$ and $d_4$) are not affected by the non-linear elasticity in bending plane, which is purely based on internal mechanical interaction ({\em i.e.} ultrasound arrays and wires in the distal bending section). We also do not include any interaction with the environment for $\phi_3$ and $d_4$ in this compensation. It is clear that the catheter shape changes due to the affect of vessel curvature on the degree of thread deflections, however, the rigidity of the ICE catheter was designed with this degree of curvature in mind. Thus, first, we assume the catheter shape is the straight to validate our method, and then We demonstrate performance changes in varying levels of vessel curvature in Section\,\ref{sec:experiments}.
}

\begin{figure}[t!]	
	\begin{center}\hspace*{-10pt}
    \includegraphics[scale= 0.65]{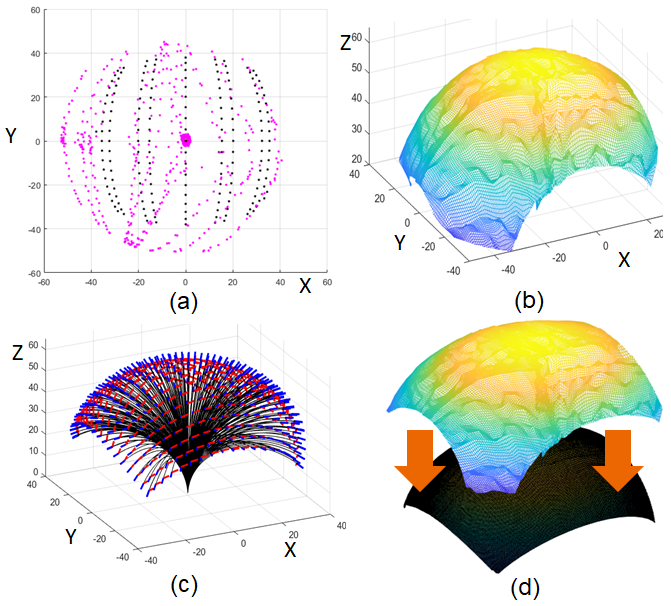}
	\end{center}\vspace*{-10pt}
	\caption{This demonstrates one example of mapping procedures. 
	The workspace is $\pm~90^{\circ}$. The bending length $L$ is $60~mm$. {\bf (a)} This shows the bending plane $X$ and $Y$ corresponding to ($\phi_1, \phi_2$); the black dots indicate the model geometry ${\bf \hat{P}}_{tip}$ from forward kinematics model in 2D projection, the purple dots indicate the real geometry ${\bf P}_{tip}$ in 2D projection, which are collected by EM sensors ($10^\circ$ intervals for $\pm 90^\circ$). The model geometry and the real geometry shows discrepancy due to non-linear torsion. {\bf(b)} This shows interpolated ${\bf P}^*$ corresponding to ($\phi_1, \phi_2$) on the $T_{base}$ coordinate. ${\bf P}^*$ is the unobserved values (i.e. the query points for interpolators). {\bf(c)} This shows the estimated position ${\bf \hat{P}}^k_{tip}$ corresponding to ($\phi_1, \phi_2$). This comes from the forward kinematics model. The red arrow indicates ${\bf \hat{n}}_{tip}^x$, while the blue arrow indicates ${\bf \hat{n}}_{tip}^z$.  {\bf(d)} We map the real geometry (b) into the model geometry (c) on $T_{base}$: both the black sphere (${\bf \hat{P}}_{tip}$) and the colored mesh (${\bf P}_{tip}$) are interpolated. Finally, the mapping function provides $(\phi'_1,\phi'_2)$ corresponding to $({\phi}_1,{\phi}_2)$.
		\label{fig:interpolation}}
	\vspace*{-10pt}
\end{figure}

Let ${\bf P_{tip}} \in \mathbb{R}^3$ be the real position $(x_{tip}, y_{tip}, z_{tip})$ of the catheter tip. When $\phi_1$ and $\phi_2$ are input to the kinematics model, the model predicted position ${\bf \hat{P}_{tip}}$ can present large discrepancies with the actual position ${\bf P_{tip}}$ due to the effects of non-linear elasticity of the bending space. We assume that the bending length of the catheter remains constant due to the arc constraints, and then the two control knobs (\textit{i.e.} anterior-posterior and right-left) cover the full working space. Accordingly, only ${\bf \hat{P}_{tip}}$ and ${\bf P_{tip}}$ are misaligned as shown in Figure\,\ref{fig:interpolation}~(a). This is an inevitable consequence based on the mechanical tolerance data shown in Figure\,\ref{fig:catheter_overall}~(c).


To increase the accuracy of the kinematics model, we propose to map the model input ($\phi_1$ and $\phi_2$ for ${\bf \hat{P}_{tip}}$) with the real joint states ($\phi'_1$ and $\phi'_2$ for ${\bf P_{tip}}$). This mapping function is applied to both forward/inverse kinematics. We define the mapping function $F$ as follows:
\begin{equation}
\phi'_{1},\phi'_{2} = F(\phi_1,\phi_2)\label{eq:interpolator}
\end{equation}

The function $F: \phi_1,\phi_2 \rightarrow \phi'_1,\phi'_2 $ is the mapping function based on correcting the estimated position, ${\bf \hat{P}_{tip}}$ to the real position values, ${\bf P_{tip}}$.
Each catheter has an individual model of elasticity behaviors within the catheter tolerance map as shown in Figure\,\ref{fig:catheter_overall}~(c).

The non-linear elasticity field learning method is addressed as follows:

First, we condition the model by collecting ground-truth data by manipulating the catheter by robot control and sampling the joint state $(\phi_1, \phi_2)$ and the real position ${\bf {P}_{tip}}$. Then, we have $S$ number of samples, which gives the real positions ${\bf {P}_{tip}}$ related to $(\phi_1, \phi_2)$. Figure\,\ref{fig:interpolation}~(a) shows an example of the sampled ground-truth, ${\bf {P}_{tip}}$, as purple dots while ${\bf \hat{P}_{tip}}$ is shown as the black dots. There exist the model errors between the black dots and the purple dots due to plastic torsion.

Second, let ${\bf {P}^*_j}=(x^*_j,y^*_j,z^*_j)$ be the unobserved values based on the whole workspace $([-d, d]~\in~\phi_1$ and $\phi_2)$, where $d$ is the maximum degree of knobs, and $j \in U$, $U$ is the number of unobserved values in the whole workspace. Then, we use a 2D interpolator with collected data to estimate  ${\bf {P}^*_j}$. We have three interpolators: $(\phi^j_1, \phi^j_2) \rightarrow x^*_j$, $(\phi^j_1, \phi^j_2) \rightarrow y^*_j$, and $(\phi^j_1, \phi^j_2) \rightarrow z^*_j$. Figure\,\ref{fig:interpolation}~(b) shows an example of interpolated ${\bf {P}^*}$. We used the cubic spline interpolant, which interpolated the value at a query point based on the values at neighboring points in two-dimension. Several methods for interpolation exist ({\em e.g.} barycentric, polynomial, spline, etc).

Third, we applied the whole workspace inputs $(\phi^k_1,\phi^k_2), k\in S+U$ into the forward kinematics model. Then, we get ${\bf \hat{P}^k_{tip}}$, which is shown in Figure\,\ref{fig:interpolation}~(c) as examples. 

Lastly, we pick the query position ${\bf \hat{P}^k_{tip}}$, and find the nearest position in ${\bf {P}}^* + {\bf P}_{tip}$. Then $(\phi^k_1,\phi^k_2) \rightarrow {\bf \hat{P}}^k_{tip}$ and ${\bf \hat{P}}^k_{tip} \rightarrow {\bf {P}}^* + {\bf P}_{tip}$, thus ${\bf {P}}^* + {\bf P}_{tip}$ give the corrected values $(\phi'_1,\phi'_2)$ corresponding to $(\phi^k_1,\phi^k_2)$. As examples, we show the overall values in Figure\,\ref{fig:interpolation}~(d).
Basically, we register the real geometry (${\bf {P}_{tip}}$) to the model geometry (${\bf \hat{P}_{tip}}$). \jc{Then, our interpolation-based mapping function represents $(\phi'_1,\phi'_2)$ corresponding to $({\phi}_1,{\phi}_2)$, which can be differentiable and invertible for kinematics back and forth.}

We demonstrate non-linear elasticity compensation in the application of rotating the ultrasound image about the catheter axis, ${\bf\hat{n}}_{tip}^z$ (z-axis of the ICE catheter), while maintaining a constant tip position. This trajectory computation is possible using inverse kinematics. One example is that: (1) Initially, we start from $(\phi_1, \phi_2) = (60^{\circ},0^{\circ})$, and rotate $360^{\circ}$ along $\phi_3$. Then, the trajectory is one cosine function of $\phi_1$, one sine function of $\phi_2$, and one negative linear function of $\phi_3$. $(\phi_1,\phi_2)$ example result is shown in Figure\,\ref{fig:phi_model}. These combined controls are challenging by hands, but robotic controls can provide easily.

\subsection{Automated View Recovery: a topological map construction and path planning}\label{sec:view-to-view}
During the procedure, we continuously construct a topological graph ({\it ``a roadmap''}) and a library of views (\textit{i.e.} important locations on the \textit{roadmap}) based on the user's trace and inputs when manipulating the catheter by joystick input to the robotic manipulator. Queries by the user to return to a specific view will be given to the controller to search a path.
More specifically, let $\mathcal{G}({\bf V},{\bf E})$ represent a graph in which {\bf V} denotes the set of configurations ${\bf q_i}$ and {\bf E} is the set of paths $({\bf{q_i}, \bf{q_j}})$. Our path planning is divided into two phases of computation: 

\vspace{5pt}
{\noindent \it Construction phase:}
The library of views and roadmap generation phase: as the robot moves, the current configuration ${\bf q_n}$ is updated. If ${\bf q_n}$ is not the same as the previous configuration ${\bf q}_{before}$,  then the algorithm inserts ${\bf q_n}$ as new vertex of $\mathcal{G}$, 
and connect pairs of ${\bf q_n}$ and existing vertices if the distance is less than a density parameter $\epsilon$ (lines\,\ref{step:vertex}-\,\ref{step:edge} of Algorithm\,\ref{algo:graph}). We apply $\epsilon$ based on Euclidean distance (assuming $1~mm \equiv 1^{\circ}$). If a larger $\epsilon$ (default = 1) were applied, then the search would be faster, however a safety of the path would not be guaranteed, as larger steps along the path could result in collision with anatomy.
Concurrent with roadmap generation, the algorithm constructs a library of views $\mathcal{V}$ when the user saves the view anytime, where $\mathcal{V} = ({{\bf q'_1},...,{\bf q'_m}})$, ${\bf q'}$ is the user saved configuration. $m$ is the number of the user saved views. This step is shown in line\,\ref{step:viewsaving} of Algorithm\,\ref{algo:graph}. 

\begin{algorithm}
	\caption{BUILD$\_$ROADMAP$\_$VIEWS~(${\bf q_n}$, $\mathcal{G}$, $\mathcal{V}$, $\epsilon$)}
	\label{algo:graph}
	\begin{algorithmic}[1]
		{
			\STATE INPUT: the current configuration ${\bf q_n}$, the current roadmap $\mathcal{G}$, the current library of views $\mathcal{V}$, the density parameter $\epsilon$
			\STATE OUTPUT: $\mathcal{G}$, $\mathcal{V}$
			\STATE Initialize: ${\bf q_{before}= [~]}$, $\mathcal{G} =[~]$, $\mathcal{V}=[~]$
			\WHILE{ROBOT is OPERATIONAL}			
			\STATE ${\bf q_n}$ is updated from the current configuration
			\IF {$\bf q_{before} \neq  q_n$}
			\FOR {each $q_i$ $\in$ NEIGHBORHOOD($q_n$,$\mathcal{G}$) }			
			\IF {$dist(q_i,q_n) \leq \epsilon$}  
			\IF {$q_n$ $\notin$ $\mathcal{G}$} 
			\STATE 	$\mathcal{G}$.add$\_$vertex($q_n$)\label{step:vertex}
			\ENDIF
			\STATE $\mathcal{G}$.add$\_$edge($q_i$,$q_n$)\label{step:edge}	
			\ENDIF	
			\ENDFOR	
			\IF {VIEW$\_$SAVING$\_$FLAG}
			\STATE $\mathcal{V}$.pushback($q_n$)\label{step:viewsaving}
			\ENDIF
			\STATE ${\bf q_{before} = q_n}$
			\ENDIF
			\ENDWHILE
		}
	\end{algorithmic}
\end{algorithm}	

\vspace{5pt}
{\noindent \it Query phase:}
Given a start configuration ${\bf q_S}$ (the current ${\bf q_n}$) and a goal configuration ${\bf q_G} \in \mathcal{V}$, is given during the procedures. Since each configuration is already in $\mathcal{G}$, we use a discrete ${A^*}$ search algorithm to obtain a sequence of edges that forms a path from ${\bf q_S}$ to ${\bf q_G}$. The more detailed roadmap construction and search algorithms in the graph are in \,\citep{lavalle06planning}.

The overall illustration of automated view recovery is shown in Figure\,\ref{fig:overview}: The roadmap to views and the library of image views are constructed in ({\it construction phase}; the blue dot shows the entire path of the user's trace during joystick manipulation of the catheter; the red dot represents the bookmarked or saved state of the robot when viewing a desired image (\textit{i.e.} corresponding to the library of views). When applied during a procedure, the user can specify the desired view (red dot). The controller then identifies a path from the current configuration to the target configuration along the edges connecting the blue dots. 



\section{Experiments and Results}\label{sec:experiments}

\subsection{Experimental Design}

\subsubsection{Data-driven compensation model }\label{sec:data-driven}

{\noindent \it Overview:~} \jc{Evaluation of the proposed compensation model was performed both on benchtop and in an abdominal aorta phantom (vascular simulations LLC) with varied curvature conditions. An EM sensor (Model 800 sensor, 3D Guidance, Northern Digital Inc.) was attached to the catheter tip to provide real-time tracking of position and orientation.}
\vspace*{5pt}

{\noindent \it Model scenarios:~} \jc{First, we collected the real position of the catheter in benchtop experiment while rastering the catheter along the full parameter space enclosed by $\pm 90^{\circ}$ in $(\phi_1,\phi_2)$ (i.e. anterior-posterior and left-right control knobs). These data were then used to evaluate the compensation model under the following conditions: (1) {\it Without compensation}, (2) {\it With compensation}, and {\it Simplified compensation}. 
\begin{enumerate}
    \item The {\it Without compensation} condition uses the full discrete set of positional data collected during this exercise to produce trajectories based on $\hat{P}$.
    \item The {\it With compensation} condition uses a dense set of $\phi_1$ and $\phi_2$ consisting of $360$ samples collected at $10^{\circ}$ intervals from the exercise to interpolate ${\bf P}^*_{tip}$ and then generate trajectories based on ${\bf P}^*_{tip}$ corresponding to $(\phi'_1,\phi'_2)$.
    \item The {\it Simplified compensation} condition similarly uses a sparse sampling of data from the workspace.
\end {enumerate}}
\vspace*{5pt}

\begin{figure}[t!]	
	\begin{center}\hspace*{-15pt}
		\subfigure[{\it Without compensation}]{\includegraphics[scale= 0.23]{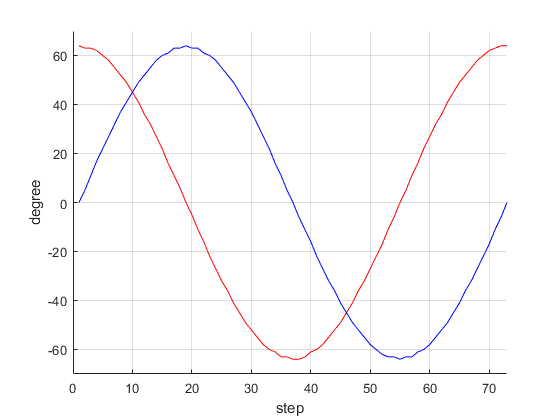}\label{fig:phi_model}}\hspace*{-20pt}
		\subfigure[{\it With compensation}]{\includegraphics[scale= 0.23]{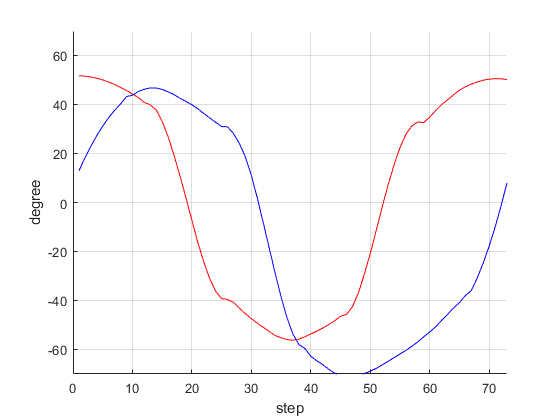}\label{fig:brute_model}}\hspace*{-20pt}				\subfigure[{\it With simplified compensation}]{\includegraphics[scale= 0.23]{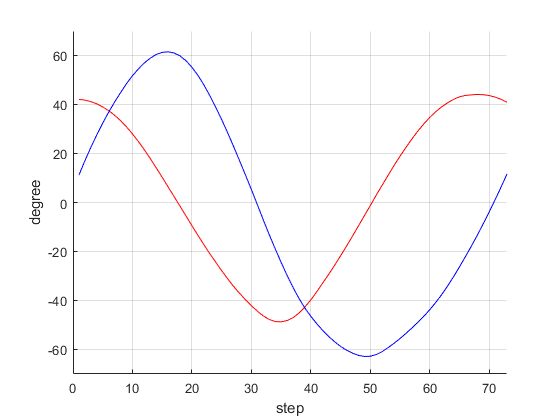}\label{fig:real_model}}\hspace*{-0pt}
	\end{center}\vspace*{-10pt}
	\caption{This is the spinning simulation. The initial setup is $(\phi_1,\phi_2)=(60^{\circ},0)$. Y-axis is degrees. X-axis is time steps.(a) This shows the base model, so without any compensation. $\phi_1$ and $\phi_2$ are intercross each other. (b) the compensated $\phi'_1$ and $\phi'_2$ are demonstrated based on the densely collected data. (c) This shows the compensated $\phi'_1$ and $\phi'_2$ based on five points samples.
		\label{fig:interpolation_error}}
	\vspace*{-15pt}
\end{figure}

{\noindent \it Model calibration:~} \jc{To achieve the sparse sampling, we propose the simple training condition based on the tolerance map in Figure\,\ref{fig:catheter_overall}~(c). From our observation, the critical point is the boundary of the arc, which is the $90^{\circ}$ from $T_{base}$ for each bending side. Therefore, we pick five points experimentally that are easy to verify visually. Practically, we manipulate two knobs using the joystick to attempt the desired positions which are determined by $(x_{tip},y_{tip})$ ({\em c.f.} $z_{tip}$ is ignored due to arc constraints.). We propose five points: two $\rm~90^{\circ}$ bending shape of the right-left planes when $x_{tip} = 0$, two $\rm~90^{\circ}$ bending shape of the anterior-posterior planes when $y_{tip} = 0$, and finally the initial position by $(x_{tip},y_{tip}) =(0,0)$. After generating the motion trajectories, we applied the Savitzky-Golay filtering method to produce a smooth path. The chosen filter smooths according to a quadratic polynomial that is fit over each windowed trajectory. Three exemplary trajectories of the testing conditions are shown in Figure\,\ref{fig:interpolation_error}.}
\vspace*{5pt}

{\noindent \it Benchtop:~} \jc{Initial evaluation of the three compensation methods was performed with the catheter and robotic controller on benchtop. The controller was tasked with revolving the ultrasound imaging plane $360^{\circ}$ about the catheter axis while maintaining a constant catheter tip position with various bending conditions applied to the catheter. This was performed from three initial conditions: a $20^{\circ}$ anterior-posterior bend, a $40^{\circ}$ anterior-posterior bend, and a $60^{\circ}$ anterior-posterior bend (\textit{i.e.} bending in $\phi_1$). {Under these conditions, the root-mean-square error (RMSE) catheter tip position and rotational error is reported by comparing the observed (EM tracking) and modelled ($\hat{P}_{tip}$) positions for the three model conditions.}}
\vspace*{5pt}

{\noindent \it Phantom:~} \jc{Phantom evaluation of the three compensation methods at various curvature conditions was performed in the abdominal aorta phantom. The front component of the robotic manipulator (Figure\,\ref{fig:overview}) was placed just proximal of the bending section of the catheter to hold the catheter firmly (\textit{i.e.} reducing the distance between the catheter tip and the manipulator to roughly 10 cm). The abdominal aorta phantom was placed along the section of the catheter between the front and rear mechanisms of the robot, allowing us to freely adjust the path of the catheter without inducing other environmental disturbances. A Tuohy borst adapter was used at the femoral access site. The abdominal aorta phantom presented two main curvatures: 1) between the inferior vena cava and the iliac vein and 2) between the iliac vein and the femoral access. We define three curvature conditions in phantom by two arcs; {\it straight} ($0^\circ$, $0^\circ$), {\it moderate} ($30^\circ$, $30^\circ$), and {\it steep} ($40^\circ$, $45^\circ$). We used three separate ICE catheters (ACUSON Acunav Volume, Siemens Healthineers) for all modeling and testing within this study.}

\subsubsection{Automated view recovery}

{\noindent \it Heart phantom study:~} \jc{Initial validation of robotic catheter control during view recovery was performed with the catheter tip inserted into the left ventricle of a custom beating heart phantom (Archetype Medical) for ICE and TEE valve imaging. An EM sensor (Model 800 sensor, 3D Guidance, Northern Digital Inc.) was attached to the catheter tip to provide real-time catheter tip tracking as the catheter was manipulated by robotic control. These EM data provide baseline evaluation of the accuracy of catheter tip positioning and orientation.}
\vspace*{5pt}

{{\noindent \it Animal study:~} Three \jc{animal} validation experiments were performed at the Houston Methodist Institute for Technology, Innovation $\&$ Education (MITIE, Houston Methodist Hospital) with the \textit{in vivo} study protocol approved by the Institutional Animal Care and Use Committee (IACUC). All testing of the robotic catheter controller was performed under general anesthesia. Vascular access was achieved bilaterally to allow for manipulation of ICE and other device catheters. \textit{In vivo} experimental setup is pictured in Figure\,\ref{fig:overview}. In each experiment, the ICE catheter, with robotic controller pre-attached, was introduced to the venous system through \jc{an} introducer sheath with balloon seal (DrySeal Flex Introducer Sheath, Gore) before being manually inserted to the junction of the inferior vena cava (IVC) and right atrium (RA) under fluoroscopic guidance. When viewing anatomy, ultrasound image data were recorded as DICOM format on an Acuson SC2000 (Siemens Healthineers). Dynamic computed tomography (DCT) images were acquired throughout experimentation on an Artis Zeego (Siemens Healthineers) to provide volumetric ground-truth catheter \jc{position} relative to anatomy ($200^{\circ}$ total rotation, $5$ second acquisition, $60$ frames-per-second).}
\vspace*{5pt}

\begin{figure}[t!]
	\centering
	\includegraphics[scale= 0.55]{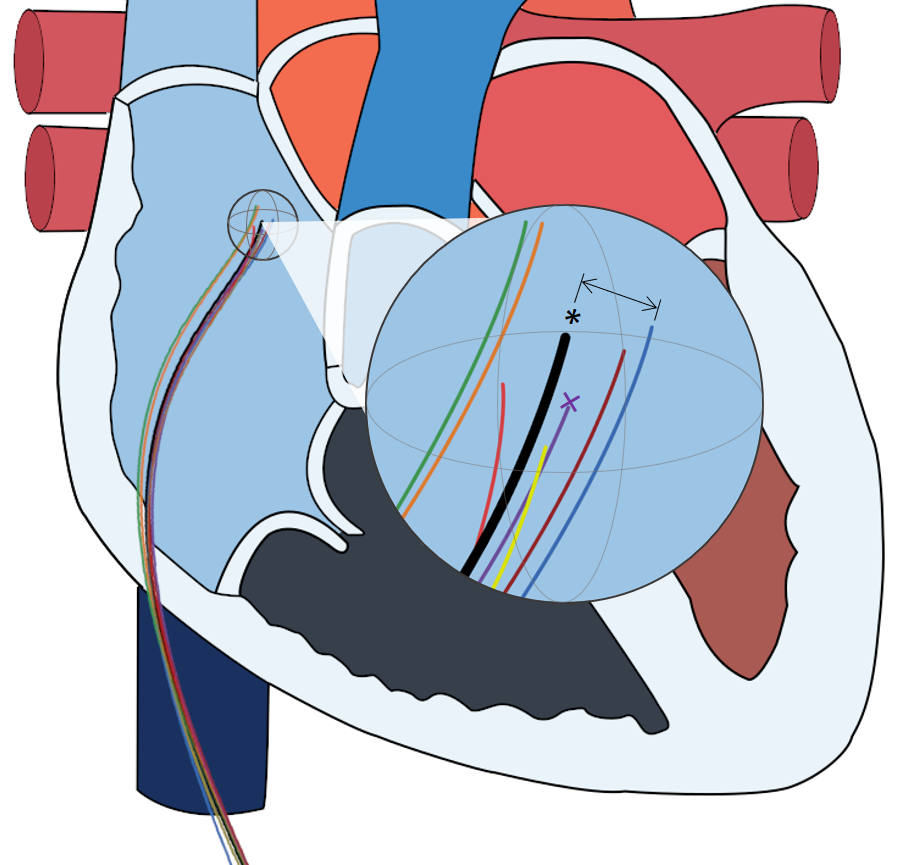}
	\vspace*{-10pt}
	\caption{Diagram of seven automated catheter returns to a single image-target. The black catheter, marked by *, represents the spatial average of the other seven. In this scenario, the purple catheter, marked by x, was designated as the reference catheter to which comparisons are made as it is the geometric median. The observed spread relative to the purple catheter represents the error associated with the automated view recovery as reported in this study. \label{fig:catheter}}
	\vspace*{-15pt}
\end{figure}

{\noindent \it Validation metrics:~} {For this study, we compare catheter tip location across multiple robotic positioning events to an image target as the primary spatial validation approach. For any catheter positioning to an image target, the robotic motors will theoretically return to an identical final state, assuming no slippage and consistent backlash. As we have created an open-loop system, when the robot executes movement along a path it is unaware of any influences external to the controller which may affect its final positioning. As such, we assume that any error in the positioning of the catheter tip are due to those external influences (\textit{e.g.} catheter-robot mechanics, intravenous catheter interactions, cardiac and respiratory motion, etc). These measurements, when taken relative to a reference catheter position, represent the distribution of uncertainty in robotic catheter positioning resulting from these external sources of error. Because of this, each experimental robotic positioning of the catheter to an image target can be considered a sampling of that uncertainty (Figure\,\ref{fig:catheter}). Therefore, we have elected to use the geometric median catheter (\textit{i.e.} the catheter which presented nearest the centroid of the sampled distribution) as the reference catheter for each imaging target when evaluating the view recovery process.

To measure \textit{in vivo}, 3D models of the ground-truth observance of each ICE catheter were generated following an intensity-based threshold segmentation of intra-procedural DCT images (voxel spacing 0.498 mm) using ITK-SNAP \,\citep{py06nimg}. Co-registration between DCT images were not necessary as no bulk subject motion was observed between scans. Next, the ICE catheter tip was manually labeled within each model and the centroid taken to define the discrete tip location for a given robotic positioning event. A curve was then fit to each catheter, constrained to terminate at the predefined catheter tip, by fitting a low-order polynomial. This process provided a discrete spatial representation of the catheter tip and body. Finally, catheter tip position error was calculated as the Euclidean distance between each catheter tip location and the respective reference catheter tip location for each imaging target.

Robotic catheter positioning was further evaluated by measuring the similarity of acquired ultrasound images between each experimental acquisition and the reference acquisition. Ultrasound image clips encompassing multiple heart cycles were acquired following each robotic positioning event. Image sequences were manually synchronized in time based on visible anatomy (\textit{e.g.} valve leaflet position; from the first frame of the target valve being open to the last frame before re-opening in the next cycle) and trimmed to encompass exactly one heartbeat. As before, image sequences from the geometric median catheter were selected as the reference images for comparison. \jc{Ultrasound images} for each series of \jc{target view} were then compared, in the \jc{labeled corresponding frames}, to the reference images by computing the image cross-correlation. In-image regions of interest (ROI) were also manually labeled for each image series \jc{(Figures\,\ref{fig:result1}(A) and \ref{fig:result2}(A,C))}. ROI centroids from each image acquisition were measured to corresponding ROI centroids in the reference images by Euclidean distance.}

\subsection{Results}
\subsubsection{\jc{Evaluation of data-driven compensation model}}

{
We addressed the results in {\it straight} condition in Table\,\ref{tab:rmse_results}. Each cell of Table\,\ref{tab:rmse_results} is represented as (tip position RMSE, tip rotational RMSE).
These results represent that the {\it Without compensation} condition provides high variance. Conversely, the {\it With compensation} condition produces relatively low variance. Similarly, the {\it Simplified compensation} condition also improves upon the {\it Without compensation} condition, providing lower variance, but to a lesser extent than the {\it With compensation} condition.}

\begin{figure}[t!]	
	\begin{center}
		\subfigure[{\it Position errors in varying curvature condition}]{\includegraphics[scale= 0.8]{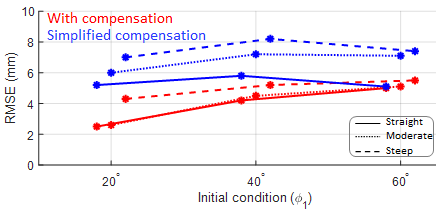}\label{fig:pos_errors}}\\
		\subfigure[{\it Orientation errors in varying curvature condition}]{\includegraphics[scale= 0.8]{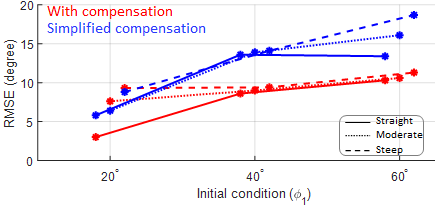}\label{fig:ori_errors}}
	\end{center}\vspace*{-10pt}
	\caption{ This demonstrates errors of our compensator in three different vasculature curvature conditions ({\it straight, moderate, steep}). The red color indicates with compensation. The blue color indicates the simplified compensation. The solid line is {\it straight}, the dotted line is {\it moderate}, and the dashed line is {\it steep} conditions.
		\label{fig:curvature_errors}}
	\vspace*{-20pt}
\end{figure}

\jc{Next, we show the error changes depending on curvature conditions in Figure\,\ref{fig:curvature_errors}. Figure\,\ref{fig:curvature_errors} shows that the curvature angle is increased as {\it straight} (solid line), {\it moderate} (dotted line), and {\it steep} (dashed line) conditions. The position RMSE is slightly increased by less than $1~mm$ for {\it With compensation}, and less than $2~mm$ for {\it Simplified compensation} in Figure\,\ref{fig:curvature_errors}(a), and the orientation RMSE is slightly increased by less than $5^\circ$ for {\it With compensation}, and less than $5^\circ$ for {\it Simplified compensation} in Figure\,\ref{fig:curvature_errors}(b). 
}

\begin{table*}[t]
\centering
\caption{\jc{RMSE between the ground-truth and estimated position/rotation during image spinning in bench-top straight condition}}
\label{tab:rmse_results}
    \begin{tabular}{|l|c|c|c|}
    \hline
    Initial condition ($\phi_1$) & $20^{\circ}$  & $40^{\circ}$ & $60^{\circ}$ \\ 
    \hline
    {\it Without compensation}     & (5.5~$mm$, 10.0$^{\circ}$)     
                & (11.0~$mm$, 20.5$^{\circ}$)          
                & (14.8~$mm$, 31.0$^{\circ}$)      \\
    {\it With compensation}     &
                (2.5~$mm$, 3.0$^{\circ}$)     
                & (4.2~$mm$, 8.6$^{\circ}$)         
                & (5.0~$mm$, 10.3$^{\circ}$)    \\
    {\it With simplified compensation} & 
                (5.2~$mm$, 5.8$^{\circ}$)     
                & (5.8~$mm$, 13.6$^{\circ}$)     
                & (5.1~$mm$, 13.4$^{\circ}$)    \\
    \hline
    \end{tabular}
\end{table*}

\subsubsection{\jc{Phantom validation of automated view recovery}}
\jc{Unique initial catheter tip positions were manipulated to within the beating heart phantom by joystick input to the robotic controller and saved (\textit{i.e.} the joystick input maps to the standard catheter control knobs, these digital inputs are then translated to the robotic motors to result in catheter manipulation). For each target, the robotic control state was then added to the library of views for later automatic return via  topological path planning. The controller was then tasked with cycling the ICE catheter tip between each target position in series a total of 16 times. Accuracy of the robotic controller was measured by an EM sensor attached at the catheter tip. The set of target positions encompassed manipulation of all 4 DOF of catheter motion. Under these conditions, the robotic controller maneuvered the catheter tip to the target position with an average position error of $0.67 \pm 0.79~mm$ and rotational error of $0.37^{\circ} \pm 0.19^{\circ}$.}

\begin{figure}[b!]
	\centering
	\includegraphics[scale= 0.355]{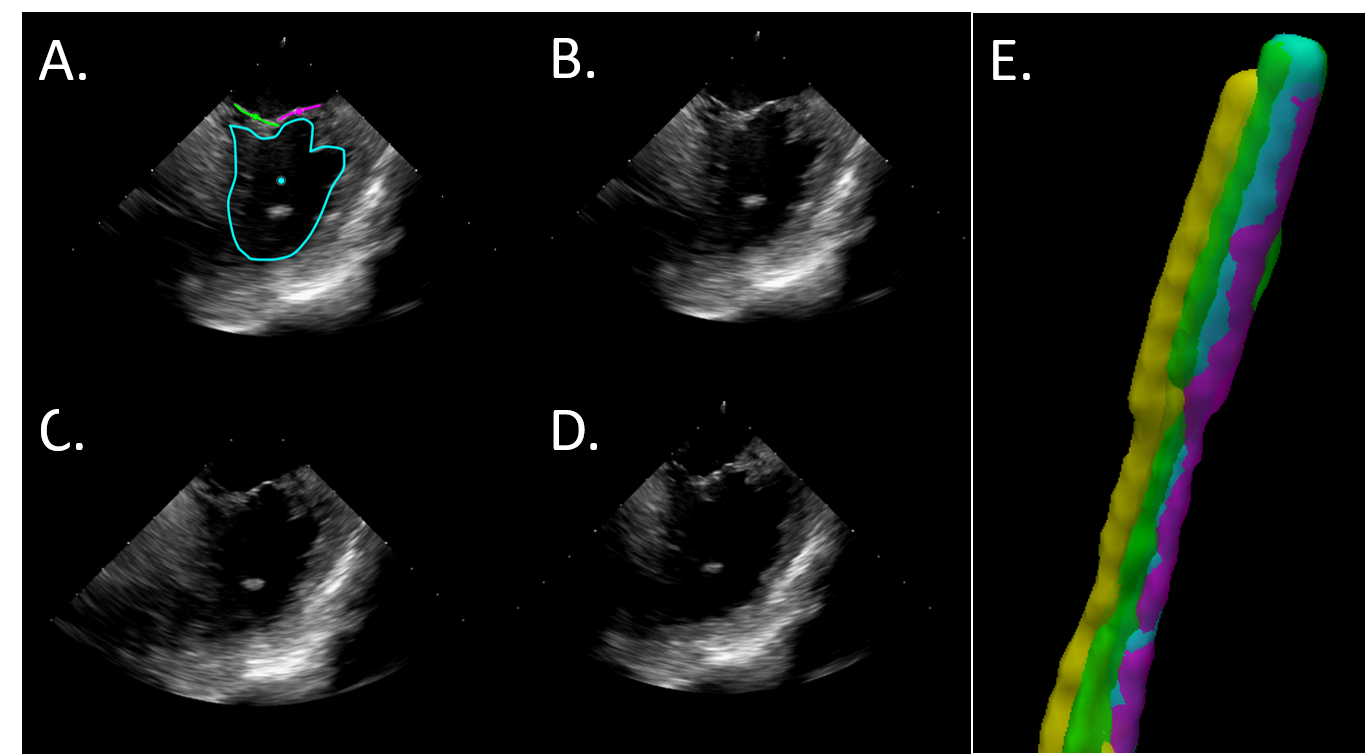}
	\vspace*{-10pt}
	\caption{Example results from {\em in vivo} validation experiments. (A) the initial view of the Tricuspid Valve from joystick control by the interventionalist. (B-D) reacquisitions by semi-autonomous robotic control of the ICE catheter. (E) presents the ground-truth catheter position of (A) in green and (B-D) in magenta, cyan, and yellow respectively. Manually labeled in-image ROI are also shown in (A). \label{fig:result1}}
	\vspace*{-0pt}
\end{figure}

\begin{figure*}[t!]
	\centering
	\includegraphics[scale= 0.425]{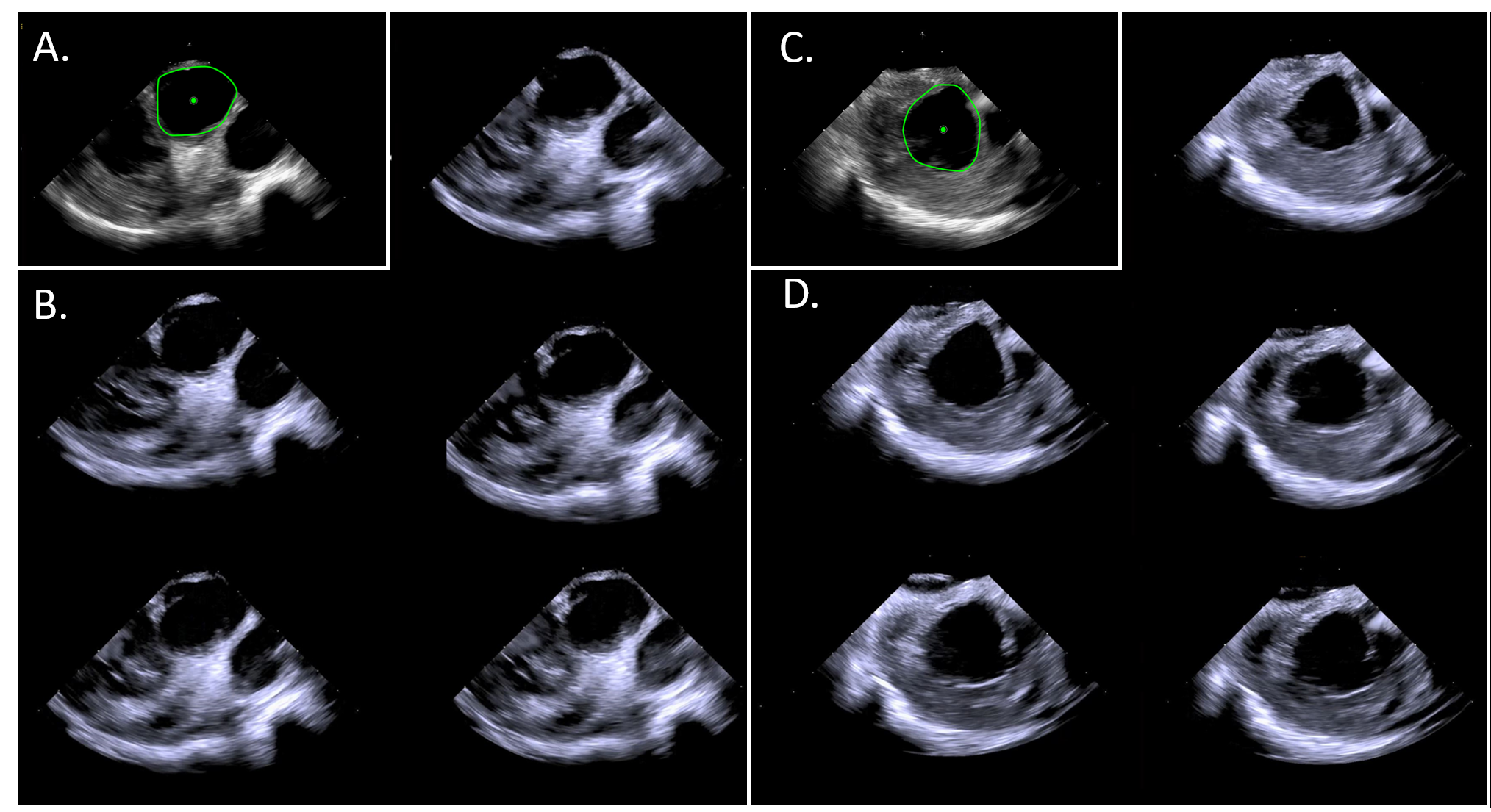}
	\vspace*{-10pt}
	\caption{(A,C) Reference and (B,D) automatically reacquired images of the Aortic and Mitral valves respectively. Manually labeled in-image ROI are shown in (A,C). \label{fig:result2}}
	\vspace*{-10pt}
\end{figure*}


\begin{table*}[h!]
\centering
\caption{Localization and image-based validation during automated view recovery}
\label{tab:table1}
\begin{tabular}{|l|cc|cc|cc|}
\hline
 &
  \multicolumn{2}{c|}{\begin{tabular}[c]{@{}c@{}}Catheter Localization [mm]\end{tabular}} &
  \multicolumn{2}{c|}{\begin{tabular}[c]{@{}c@{}}Image Similarity [\%]\end{tabular}} &
  \multicolumn{2}{c|}{\begin{tabular}[c]{@{}c@{}}ROI Localization  [mm]\end{tabular}} \\ \cline{2-7} 
Image Target    & $\mu$ & $\sigma$ & $\mu$  & $\sigma$ & $\mu$ & $\sigma$ \\ \hline
Aortic Valve    & 1.4 & 1.1    & 81.4 & 5.6    & 2.8 & 1.4    \\
Mitral Valve    & 1.6 & 0.9    & 80.9 & 5.6    & 1.8 & 0.8    \\
Tricuspid Valve & 1.4 & 0.8    & 81.0 & 9.3    & 4.0 & 3.0    \\ \hline
\end{tabular}
\end{table*}

\subsubsection{\jc{Animal validation of automated view recovery, position analysis}}
Initial ultrasound views of the Aortic, Mitral, and Tricuspid valve imaging targets were manipulated \jc{to and saved} by the physician using joystick input to the robotic controller. Views were then automatically returned to in series several times by automated robotic control (\textit{e.g.} consecutively cycling through return to Aortic \jc{valve}, return to Mitral \jc{valve}, return to Tricuspid \jc{valve}). The ultrasound image was recorded for several heart cycles and a DCT image was acquired for each viewing of an anatomical target. These data were then compared to measure the reproducibility of robot controlled catheter \jc{tip} motions. In total, three independent \jc{animal} experiments evaluating automated view recovery were performed with four, six, and seven imaging events respectively per image target, resulting in 51 total view recoveries.

In animal catheter tip localization error from DCT is presented in Table \ref{tab:table1} for each imaging target. Across all image-targets, average catheter tip localization error was $1.5 \pm 0.9$ mm.


\subsubsection{\jc{Animal validation of automated view recovery, image analysis}}
Example images from independent robotic re-acquisitions of the Tricuspid valve are presented in Figure\,\ref{fig:result1} alongside the corresponding ground-truth catheter segmentations from DCT. Further examples of the Aortic and Mitral valves are presented in Figure\,\ref{fig:result2}. Image similarity as measured by image cross-correlation is presented in Table \ref{tab:table1} for each image target. Average image similarity across all image-targets and experiments was $81.1 \pm 6.9~\%$. For comparison and to establish \jc{a relative maximum achievable similarity}, the average image similarity when comparing images from immediately consecutive heart cycles (\textit{i.e.} without moving the catheter) was $89.8~\%$. \jc{This value represents the an aggregate of the observable noise resulting from slight variations in cardiac rhythm, respiratory motion, and ultrasound speckle.} 


Finally, in-image ROI were designated for each image target based on \jc{visible} anatomy (Figures\,\ref{fig:result1}(A) and \ref{fig:result2}(A,C)). ROI centroids were then measured relative to corresponding reference ROI to quantify \jc{the localization of ROI relative to the imager}. Average in-image ROI localization error across all image-targets was $2.9 \pm 2.2$. Again for comparison and to establish an accuracy floor, average ROI localization error when comparing ROI from immediately consecutive heart cycles (\textit{i.e.} without moving the catheter) was $2.0$ mm.


\section{Discussion}



The methods presented in this study represent the first work to demonstrate automated image recovery through robotic ICE catheter control. While there are several commercially available robotic catheter control systems on the market, there is only one commercial system, Vdrive (Stereotaxis Inc., St. Louis, MO, USA), which can provide limited manipulation of ICE catheters and one research prototype with a fully articulated 4 DOF \,\citep{loschak2016rob}. In \,\citet{brattain2014instrument}, robotic methods were introduced for stitching together volumetric images of an ROI from 2D ICE images and tracking of a device catheter within images. Similarly, our work further advances ICE catheter robotics by augmenting the robotic control scheme with a spatially aware process to achieve more autonomous imaging; therefore, streamlining the user experience and allowing the interventionalist to not divide focus throughout the procedure in order to image.

In this work, we apply a robotic ICE controller to implement a method for constructing a case-specific library of desired views and achieve automated recovery of those views during the procedure. In practice, standardized anatomical imaging views are emerging as indications for ICE continue to expand across disciplines. \citet{enriquez2018use} detail standard imaging views for a variety of structural and electrophysiology procedures. Similarly, early identification and unobstructed monitoring of potentially life-threatening complications is one of the most valuable functions that ICE provides \,\citep{hijazi2001asd, filgueiras2015abl, kim2009use, banchs2010intracardiac} and is leading to its adoption as a primary imaging modality for certain procedures, especially those involving transseptal puncture for left heart catheterization \,\citep{silvestry2009ice, bazaz2003site}. In response, we have developed the method presented in this paper to enable fast and precise cycling through selected views to both observe therapeutic delivery and monitor for procedural complications. If implemented in practice, these methods provide the interventionalist a means to initialize desired views covering the treatment area and then automatically return to those views at any point during a procedure. We have validated the accuracy of this approach in a series of animal experiments by a combination of spatial analysis (Figure\,\ref{fig:catheter}) and image similarity measures (Figures\,\ref{fig:result1}\,-\,\ref{fig:result2}).

Relative to typical right atrial measurements from echocardiography ($3.6 \pm 0.1$ cm and $4.2 \pm 0.1$ cm for short- and long- axis respectively \citep{bommer1979determination}), the level of catheter tip placement error observed in the automated view recovery \textit{in vivo} experiments by DCT (\textit{i.e.} $1.5 \pm 0.9$ mm) demonstrate high accuracy of by robotic catheter manipulations. These results are further corroborated by the \jc{phantom} catheter tip localization results acquired by EM sensing (\textit{i.e.} $0.67$ mm and $0.37^{\circ}$). However, we must note that the measurements taken \jc{in phantom} are below the reported positional accuracy \jc{for localization}(\textit{i.e.} $1.4$ mm) \jc{and orientation} (\textit{i.e.} $0.5^{\circ}$)) and should be considered as such. In contrast, the \textit{in vivo} validation measurements are well within the resolution of the DCT imaging. Comparing these \textit{in vivo} and \jc{phantom} results demonstrates a compounding of error sources when moving between experimental settings which include: mechanical coupling of the catheter, robot, and sheath; interactions between the catheter and vascular anatomy; and physiological sources such as cardiac or respiratory motion. When considering the multiple sources of error and how they manifest in the catheter repositioning, it highlights a major limitation of our \textit{in vivo} study design. Error accumulation affects not only the catheter tip position but also the orientation, or heading, of the ultrasound imaging plane. While we report high accuracy in tip positioning, we were unable to discretely measure misalignment of the imaging plane \textit{in vivo}. As we see an increase in the positional error when moving from \jc{phantom} to \textit{in vivo} settings ({\em i.e.} 0.67 mm to 1.5 mm), we would expect an increase in orientation error of \jc{at least} similar magnitude. As a surrogate measurement, we present the image-based validation approaches. Figures\,\ref{fig:result1} and \ref{fig:result2} along with the similarity and ROI results in Table \ref{tab:table1} represent a high degree of 2D spatial and content similarity in the produced images following robotic catheter manipulation. \jc{We further note that the selected cross-sectional views for each anatomical target were wholly visible in all acquired \textit{in vivo} ultrasound data. While the presented results focus on 2D ultrasound imaging, the experiments were performed with robotically controlled volumetric ICE. Therefore, the interventionalist can theoretically reformat the volumetric ultrasound data in order to correct for any differences in orientation without needing to adjust the catheter.} Altogether, we believe that these results support our conclusion that the robotic controller and automated view recovery process introduced in this study can accurately position the ICE catheter within the RA and reproducibly image cardiac structures of clinical relevance. However, in future \textit{in vivo} validation work, ground-truth validation of imager orientation should be collected by some means (\textit{e.g.} EM sensing, measuring to anatomical targets in volumetric ICE image-data, or \jc{contrast-enhanced} high resolution cardiac-gated cross-sectional imaging such as CT).

Safety is a major concern when considering any degree of robotic autonomy in human interfacing applications. As described, the automated view recovery method is not autonomously seeing views in an active manner, but rather relying on the user's initialization of each view and their tracing of possible paths (\textit{i.e.} roadmap $\mathcal{G}$). As the user controls the ICE catheter during the procedure, they may cause the catheter to contact critical anatomies (\textit{e.g.} septal wall, valves). However, the ICE catheter's end portion and tip are designed to be yielding so that the catheter will not puncture or damage tissue. With our presented semi-autonomous functionality, we rely on the user's previous navigation which we assume to conform with standard clinical practices. Furthermore, our spatial- and image-based results support that the robotic controller can maintain accurate and safe accordance with the expert user's trace.

It is also important that we discuss the magnitude of observed errors in Table\,\ref{tab:rmse_results} and Table\,\ref{tab:table1} and to clarify the underlying studies.

\vspace{5pt}
\jc{\bf \noindent Table\,\ref{tab:rmse_results} error discussion:}
Table\,\ref{tab:rmse_results} presents results from image spinning in the benchtop and phantom setting with applied nonlinear elasticity compensation using the \textit{Without compensation}, \textit{With compensation}, and \textit{Simplified compensation} conditions. From the results in Table\,\ref{tab:rmse_results}, both the \textit{With compensation} and \textit{Simplified compensation} conditions reduce model error. \jc{However, errors still exist in the densely sampled, \textit{With compensation} method. 
As we mentioned in section\,\ref{sec:icecatheter}, there exist dead-zone, elasticity, and hysteresis in the two pairs of tendons to drive the catheter, which are not mechanically separable. However, in this paper we focused on non-linear elasticity behavior in the bending section.
Thus, we hypothesize that these observed errors are due to dead-zone and hysteresis which are not fully accounted for by the compensation model. In future works, it may be important to integrate hysteresis modeling ({\em e.g.,}\,\citep{xu17tendon,wang20hysteresis}) alongside the proposed compensation method.}

However, dead-zone may be treated in our framework. We can observe one interesting compensation in Figure\,\ref{fig:brute_model}. This represents an example of a trajectory from the {\it With compensation} condition. Here a relatively rough trajectory was observed even after filtering was applied. Specifically, rather steep slopes exist near the zero position, which may be able to compensate for dead-zone. As expected, this was not visible in the {\it Simplified compensation} condition due to the sparse sampling of training data.

\vspace{5pt}
\jc{\bf \noindent Table\,\ref{tab:table1} error discussion:}
Table\,\ref{tab:table1} presents \textit{in vivo} validation of the automated view recovery approach. As described in Section \ref{sec:view-to-view}, this method does not rely on or apply the compensation method detailed in this work. View recovery is achieved by topological mapping of each motor state during joystick-defined manipulation of the ICE catheter by the user and subsequent pathfinding within the \jc{constructed} topological map to a desired image. In their current state, each method is applied in an open-loop scenario and represent independent initial steps towards an eventual closed-loop or more autonomous navigation approach for ICE imaging.

\section{Conclusion}
While ICE has many benefits when compared to other imaging modalities, it has limitations for clinical use largely due to difficulties in learning to control, produce, and interpret the ultrasound image.
It is clear from these limitations that a robotic-assist system to oversee significant portions of procedural ICE usage could ease the workload and burden of the interventionalist, enable complex procedures that are currently impractical for vascular intervention, and reduce the overall resource burden of procedures by easing the transition from TEE to ICE imaging.  
We suggest that the field can be further advanced, and the driving clinical need better addressed, by augmenting the robotic controller with spatial- and image-based applications that provide direct input to the motor control loop and achieve more autonomous ICE imaging. Moreover, in our review of clinical literature, we have identified several natural use cases in standard ICE imaging that are synergistic with the application of advanced robotic control.

We believe that this work represents an advancement in the application of robotics for the challenging environment of intracardiac imaging. Herein, we present and validate in \jc{phantom and} animal the first semi-autonomous ultrasound image recovery method for automating clinical ICE imaging in a natural use case. This work demonstrates that robotic control can be applied to accurately and reproducibly image in a systematic fashion. While further investigation is required to fully characterize and compensate the effects of various sources of error on both catheter kinematics modeling and \textit{in vivo} catheter control, the evaluated methods appear promising. Based on this work, we recommend that the field of intracardiac imaging continue developing towards procedural automation and standardization which can be enabled through robotic assistance. We believe these data are supportive that robotic control can be reliably applied to automate standard processes within the clinical workflow.

\section*{Disclaimer}
{

The concepts and information presented in this abstract/paper are based on research results that are not commercially available. Future availability cannot be guaranteed.

%




}
\vspace*{0pt}
{
\small
\bibliography{references_tmrb}

\begin{thebibliography}{58}
\providecommand{\natexlab}[1]{#1}
\providecommand{\url}[1]{#1}
\csname url@samestyle\endcsname
\providecommand{\newblock}{\relax}
\providecommand{\bibinfo}[2]{#2}
\providecommand{\BIBentrySTDinterwordspacing}{\spaceskip=0pt\relax}
\providecommand{\BIBentryALTinterwordstretchfactor}{4}
\providecommand{\BIBentryALTinterwordspacing}{\spaceskip=\fontdimen2\font plus
\BIBentryALTinterwordstretchfactor\fontdimen3\font minus
  \fontdimen4\font\relax}
\providecommand{\BIBforeignlanguage}[2]{{%
\expandafter\ifx\csname l@#1\endcsname\relax
\typeout{** WARNING: IEEEtranN.bst: No hyphenation pattern has been}%
\typeout{** loaded for the language `#1'. Using the pattern for}%
\typeout{** the default language instead.}%
\else
\language=\csname l@#1\endcsname
\fi
#2}}
\providecommand{\BIBdecl}{\relax}
\BIBdecl

\bibitem[Epstein et~al.(1998)Epstein, Smith, and TenHoff]{epstein1998ep}
L.~Epstein, T.~Smith, and H.~TenHoff, ``Nonfluoroscopic transseptal
  catheterization: safety and efficacy of intracardiac echocardiographic
  guidance,'' \emph{Journal of Cardiovascular Electrophysiology}, vol.~9,
  no.~6, pp. 625--630, 1998.

\bibitem[Daoud et~al.(1999)Daoud, Kalbfleisch, and Hummel]{daoud1999ep}
E.~Daoud, S.~Kalbfleisch, and J.~Hummel, ``Intracardiac echocardiography to
  guide transseptal left heart catheterization for radiofrequency catheter
  ablation,'' \emph{Journal of Cardiovascular Electrophysiology}, vol.~10,
  no.~3, pp. 358--363, 1999.

\bibitem[Calo et~al.(2002)Calo, Lamberti, Loricchio, D'Alto, Castro, Boggi, and
  et~al.]{calo2002ep}
L.~Calo, F.~Lamberti, M.~Loricchio, M.~D'Alto, A.~Castro, A.~Boggi, and et~al.,
  ``Intracardiac echocardiography: from electroanatomic correlation to clinical
  application in interventional electrophysiology,'' \emph{Italian Heart
  Journal}, vol.~3, no.~7, pp. 387--398, 2002.

\bibitem[Basman et~al.(2017)Basman, Parmar, and Kronzon]{basman2017shd}
C.~Basman, Y.~Parmar, and I.~Kronzon, ``Intracardiac echocardiography for
  structural heart and electrophysiological interventions,'' \emph{Current
  Cardiology Reports}, vol.~19, no.~10, p. 102, 2017.

\bibitem[Rigatelli(2005)]{rigatelli2005chd}
G.~Rigatelli, ``Expanding the use of intracardiac echocardiography in
  congenital heart disease catheter-based interventions,'' \emph{Journal of the
  American Society of Echocardiography}, vol.~18, pp. 1230--1231, 2005.

\bibitem[Tan and Aboulhosn(2019)]{tan2019chd}
W.~Tan and J.~Aboulhosn, ``Echocardiographic guidance of interventions in
  adults with congenital heart defects,'' \emph{Cardiovascular Diagnosis and
  Therapy}, vol.~9, no.~2, pp. S346--S359, 2019.

\bibitem[Silvestry et~al.(2009)Silvestry, Kerber, Brook, Carroll, Eberman,
  Goldstein, and et~al.]{silvestry2009ice}
F.~Silvestry, R.~Kerber, M.~Brook, J.~Carroll, K.~Eberman, S.~Goldstein, and
  et~al., ``Echocardiography-guided interventions,'' \emph{Journal of the
  American Society of Echocardiography}, vol.~22, no.~3, pp. 213--231, 2009.

\bibitem[Green et~al.(2004)Green, Hansgen, and Carroll]{green2004mv}
N.~Green, A.~Hansgen, and J.~Carroll, ``Initial clinical experience with
  intracardiac echocardiography in guiding balloon mitral valvuloplasty:
  technique, safety, utility, and limitations,'' \emph{Catheterization and
  Cardiovascular Interventions}, vol.~63, no.~3, pp. 385--394, 2004.

\bibitem[Bartel et~al.(2011)Bartel, Bonaros, Muller, Friedrich, Grimm,
  Velik-Salchner, and et~al]{bartel2011tavi}
T.~Bartel, N.~Bonaros, L.~Muller, G.~Friedrich, M.~Grimm, C.~Velik-Salchner,
  and et~al, ``Intracardiac echocardiography: a new guiding tool for
  transcatheter aortic valve replacement,'' \emph{Journal of the American
  Society of Echocardiography}, vol.~24, pp. 966--975, 2011.

\bibitem[Ahmari et~al.(2012)Ahmari, Amro, Otabi, Abdullah, Kasab, and
  Amri]{ahmari2012mv}
S.~Ahmari, A.~Amro, M.~Otabi, M.~Abdullah, S.~Kasab, and H.~Amri, ``Initial
  experience of using intracardiac echocardiography (ice) for guiding balloon
  mitral valvuloplasty (bmv),'' \emph{Journal of the Saudi Heart Association},
  vol.~24, no.~1, pp. 23--27, 2012.

\bibitem[Marmagkiolis and Cilingiroglu(2013)]{marmagkiolis2013mv}
K.~Marmagkiolis and M.~Cilingiroglu, ``Intracardiac echocardiography guided
  percutaneous mitral balloon valvuloplasty,'' \emph{Revista Portuguesa de
  Cardiologia}, vol.~32, pp. 337--339, 2013.

\bibitem[Henning et~al.(2014)Henning, Mueller, Mueller, Zuern, Walker, Gawaz,
  and et~al.]{henning2014mv}
A.~Henning, I.~Mueller, K.~Mueller, C.~Zuern, T.~Walker, M.~Gawaz, and et~al.,
  ``Percutaneous edge-to-edge mitral valve repair escorted by left atrial
  intracardiac echocardiography (ice),'' \emph{Circulation}, vol. 130, no.~20,
  pp. e173--e174, 2014.

\bibitem[Bartel et~al.(2016)Bartel, Edris, Velik-Salchner, and
  Muller]{bartel2016tavi}
T.~Bartel, A.~Edris, C.~Velik-Salchner, and S.~Muller, ``Intracardiac
  echocardiography for guidance of transcatheter aortic valve implantation
  under monitored sedation: a solution to a dilemma?'' \emph{European Heart
  Journal of Cardiovascular Imaging}, vol.~17, no.~1, pp. 1--8, 2016.

\bibitem[Saji et~al.(2016)Saji, Rossi, Ailawadi, Dent, Ragosta, and
  Lim]{saji2016mv}
M.~Saji, A.~Rossi, G.~Ailawadi, J.~Dent, M.~Ragosta, and D.~Lim, ``Adjunctive
  intracardiac echocardiography imaging from the left ventricle to guide
  percutaneous mitral valve repair with the mitraclip in patients with failed
  prior surgical rings,'' \emph{Catheterization and Cardiovascular
  Interventions}, vol.~87, no.~2, pp. e75--e82, 2016.

\bibitem[Patzelt et~al.(2017)Patzelt, Schreieck, Camus, and
  et~al.]{patzelt2017mv}
J.~Patzelt, J.~Schreieck, E.~Camus, and et~al., ``Percutaneous mitral valve
  edge-to-edge repair using volume intracardiac echocardiography—first in
  human experience,'' \emph{Cardiovasc Imag Case Report}, vol.~1, no.~1, pp.
  41--43, 2017.

\bibitem[Rao et~al.(2005)Rao, Saksena, and Mitruka]{rao2005laa}
H.~Rao, S.~Saksena, and R.~Mitruka, ``Intra-cardiac echocardiography guided
  cardioversion to help interventional procedures (ice-chip) study: Study
  design and methods,'' \emph{Journal of Interventional Cardiac
  Electrophysiology}, vol.~13, pp. 31--36, 2005.

\bibitem[Shah et~al.(2008)Shah, Bardo, Sugeng, Weinert, Lodato, Knight, and
  et~al.]{shah2008laa}
S.~Shah, D.~Bardo, L.~Sugeng, L.~Weinert, J.~Lodato, B.~Knight, and et~al.,
  ``Real-time three-dimensional transesophageal echocardiography of the left
  atrial appendage: initial experience in the clinical setting,'' \emph{Journal
  of the American Society of Echocardiography}, vol.~21, no.~12, pp.
  1362--1368, 2008.

\bibitem[Ren et~al.(2013)Ren, Marchlinksy, Supple, Hutchinson, Garcia, Riley,
  and et~al.]{ren2013laa}
J.~Ren, F.~Marchlinksy, G.~Supple, M.~Hutchinson, F.~Garcia, M.~Riley, and
  et~al., ``Intracardiac echocardiographic diagnosis of thrombus formation in
  the left atrial appendage: a complementary role to transesophageal
  echocardiography,'' \emph{Echocardiography}, vol.~30, pp. 72--80, 2013.

\bibitem[Anter et~al.(2014)Anter, Silverstein, Tschabrunn, Schvilkin, Haffajee,
  Zimetbaum, and et~al.]{anter2014laa}
E.~Anter, J.~Silverstein, C.~Tschabrunn, A.~Schvilkin, C.~Haffajee,
  P.~Zimetbaum, and et~al., ``Comparison of intracardiac echocardiography and
  transesophageal echocardiography for imaging of the right and left atrial
  appendages,'' \emph{Heart Rhythm}, vol.~11, no.~11, pp. 1890--1897, 2014.

\bibitem[Berti et~al.(2014)Berti, Paradossi, Meucci, Trianni, Tzikas, Rezzaghi,
  and et~al.]{berti2014laa}
S.~Berti, U.~Paradossi, F.~Meucci, G.~Trianni, A.~Tzikas, M.~Rezzaghi, and
  et~al., ``Periprocedural intracardiac echocardiography for left atrial
  appendage closure: a dual-center experience,'' \emph{JACC Cardiovascular
  Interventions}, vol.~7, pp. 1036--1044, 2014.

\bibitem[Matsuo et~al.(2016)Matsuo, Neuzil, Petru, Chovanec, Janotka, Choudry,
  and et~al.]{matsuo2016laa}
Y.~Matsuo, P.~Neuzil, J.~Petru, M.~Chovanec, M.~Janotka, S.~Choudry, and
  et~al., ``Left atrial appendage closure under intracardiac echocardiographic
  guidance: feasibility and comparison with transesophageal echocardiography,''
  \emph{Journal of the American Heart Association}, vol.~5, no.~10, p.~4, 2016.

\bibitem[Hijazi et~al.(2001)Hijazi, Wang, Cao, Koenig, Waight, and
  Lang]{hijazi2001asd}
Z.~Hijazi, Z.~Wang, Q.~Cao, P.~Koenig, D.~Waight, and R.~Lang, ``Transcatheter
  closure of atrial septal defects and patent foramen ovale under intracardiac
  echocardiographic guidance: feasibility and comparison with transesophageal
  echocardiography,'' \emph{Catheterization and Cardiovascular Interventions},
  vol.~52, no.~2, pp. 194--199, 2001.

\bibitem[Mullen et~al.(2003)Mullen, Dias, Walker, Siu, Benson, and
  McLaughlin]{mullen2003asd}
M.~Mullen, B.~Dias, F.~Walker, S.~Siu, L.~Benson, and P.~McLaughlin,
  ``Intracardiac echocardiography guided device closure of atrial septal
  defects,'' \emph{Journal of the American College of Cardiology}, vol.~41, pp.
  285--292, 2003.

\bibitem[Medford et~al.(2014)Medford, Taggart, Cabalka, Cetta, Reeder, Hagler,
  and et~al.]{medford2014asd}
B.~Medford, N.~Taggart, A.~Cabalka, F.~Cetta, G.~Reeder, D.~Hagler, and et~al.,
  ``Intracardiac echocardiography during atrial septal defect and patent
  foramen ovale device closure in pediatric and adolescent patients,''
  \emph{Journal of the American Society of Echocardiography}, vol.~27, no.~9,
  pp. 984--990, 2014.

\bibitem[Kalman et~al.(1997)Kalman, Fitzpatrick, Olgin, and
  et~al.]{kalman1997abl}
J.~Kalman, A.~Fitzpatrick, J.~Olgin, and et~al., ``Biophysical characteristics
  of radiofrequency lesion formation in vivo : dynamics of catheter
  tip–tissue contact evaluated by intracardiac echocardiography,''
  \emph{American Heart Journal}, vol. 133, pp. 8--18, 1997.

\bibitem[Saliba and Thomas(2008)]{saliba2008abl}
W.~Saliba and J.~Thomas, ``Intracardiac echocardiography during catheter
  ablation of atrial fibrillation,'' \emph{Europace}, vol.~10, no.~3, pp.
  42--47, 2008.

\bibitem[Bhatia et~al.(2010)Bhatia, Humphries, Chandrasekaran, and
  Srivathsan]{bhatia2010abl}
N.~Bhatia, J.~Humphries, K.~Chandrasekaran, and K.~Srivathsan, ``Atrial
  fibrillation ablation in cor triatriatum: value of intracardiac
  echocardiography,'' \emph{Journal of Interventional Cardiac
  Electrophysiology}, vol.~28, no.~2, pp. 153--155, 2010.

\bibitem[Filgueiras-Rama et~al.(2015)Filgueiras-Rama, de~Torres-Alba,
  Castrejon-Castrejon, Estrada, Figueroa, Salvador-Montanes, and
  et~al.]{filgueiras2015abl}
D.~Filgueiras-Rama, F.~de~Torres-Alba, S.~Castrejon-Castrejon, A.~Estrada,
  J.~Figueroa, O.~Salvador-Montanes, and et~al., ``Utility of intracardiac
  echocardiography for catheter ablation of complex cardiac arrhythmias in a
  medium-volume training center,'' \emph{Echocardiography}, vol.~32, no.~4, pp.
  660--670, 2015.

\bibitem[Vitulano et~al.(2015)Vitulano, Pazzano, Pelargonio, and
  et~al.]{vitulano2015ice}
N.~Vitulano, V.~Pazzano, G.~Pelargonio, and et~al., ``Technology update:
  intracardiac echocardiography - a review of the literature,'' \emph{Medical
  Devices (Auckland, NZ)}, vol.~8, pp. 231--239, 2015.

\bibitem[Stereotaxis(2020)]{stereotaxis20}
Stereotaxis, ``Stereotaxis v-drive robotic navigation system,'' Feb. 2020,
  {http://www.stereotaxis.com/products/vdrive}.

\bibitem[{Loschak} et~al.(2017){Loschak}, {Brattain}, and {Howe}]{loschak16ice}
P.~M. {Loschak}, L.~J. {Brattain}, and R.~D. {Howe}, ``Algorithms for
  automatically pointing ultrasound imaging catheters,'' \emph{IEEE
  Transactions on Robotics}, vol.~33, no.~1, pp. 81--91, Feb 2017.

\bibitem[Loschak et~al.(2017)Loschak, Degirmenci, and
  Howe]{loschak17predictive}
P.~M. Loschak, A.~Degirmenci, and R.~D. Howe, ``{Predictive Filtering in Motion
  Compensation with Steerable Cardiac Catehters},'' in \emph{Proceedings of the
  International Conference on Robotics and Automation}, Singapore, Singapore,
  May 2017.

\bibitem[Robert J. Webster~III(2000)]{webster10ccc}
B.~A.~J. Robert J. Webster~III, ``Design and kinematic modeling of constant
  curvature continuum robots: A review,'' \emph{International Journal of
  Robotics Research}, vol.~29, no.~13, pp. 1661--1683, 2000.

\bibitem[{Ott} et~al.(2011){Ott}, {Nageotte}, {Zanne}, and {de
  Mathelin}]{ott11endoscopy}
L.~{Ott}, F.~{Nageotte}, P.~{Zanne}, and M.~{de Mathelin}, ``Robotic assistance
  to flexible endoscopy by physiological-motion tracking,'' \emph{IEEE
  Transactions on Robotics}, vol.~27, no.~2, pp. 346--359, 2011.

\bibitem[Kato et~al.(2014)Kato, Okumura, Kose, Takagi, and Hata]{kao14tendon}
T.~Kato, I.~Okumura, H.~Kose, K.~Takagi, and N.~Hata, ``{Extended Kinematic
  Mapping of Tendon-Driven Continuum Robot for Neuroendoscopy},'' in
  \emph{Proceedings of the International Conference on Intelligent Robots and
  Systems}, Chicago, USA, Sep. 2014.

\bibitem[Do et~al.(2014)Do, Tjahjowidodo, Lau, Yamamoto, and
  Phee]{do14hysteresis}
T.~Do, T.~Tjahjowidodo, M.~Lau, T.~Yamamoto, and S.~Phee, ``Hysteresis modeling
  and position control of tendon-sheath mechanism in flexible endoscopic
  systems,'' \emph{Mechatronics}, vol.~24, no.~1, pp. 12 -- 22, 2014.

\bibitem[Xu et~al.(2017)Xu, Poon, Yam, and Chiu]{xu17tendon}
W.~Xu, C.~C.~Y. Poon, Y.~Yam, and P.~W.~Y. Chiu, ``Motion compensated
  controller for a tendon-sheath-driven flexible endoscopic robot,'' \emph{The
  International Journal of Medical Robotics and Computer Assisted Surgery},
  vol.~13, no.~1, p. e1747, 2017.

\bibitem[Chen et~al.(2006)Chen, Redarce, and Redarce]{chen06kinematics}
G.~Chen, M.~T. Redarce, and T.~Redarce, ``{Development and kinematic analysis
  of a silicone-rubber bending tip for colonoscopy},'' in \emph{Proceedings of
  the International Conference on Intelligent Robots and Systems}, Beijing,
  China, Oct. 2006.

\bibitem[Xu and Simaan(2008)]{kai08continuum}
K.~Xu and N.~Simaan, ``An investigation of the intrinsic force sensing
  capabilities of continuum robots,'' \emph{IEEE Transactions on Robotics},
  vol.~24, no.~3, pp. 576--587, 2008.

\bibitem[Simaan et~al.(2009)Simaan, Xu, Wei, Kapoor, Kazanzides, Taylor, and
  Flint]{simaan10design}
N.~Simaan, K.~Xu, W.~Wei, A.~Kapoor, P.~Kazanzides, R.~Taylor, and P.~Flint,
  ``Design and integration of a telerobotic system for minimally invasive
  surgery of the throat,'' \emph{The International Journal of Robotics
  Research}, vol.~28, no.~9, pp. 1134--1153, 2009.

\bibitem[{Priester} et~al.(2013){Priester}, {Natarajan}, and
  {Culjat}]{priester13us}
A.~M. {Priester}, S.~{Natarajan}, and M.~O. {Culjat}, ``Robotic ultrasound
  systems in medicine,'' \emph{IEEE Transactions on Ultrasonics,
  Ferroelectrics, and Frequency Control}, vol.~60, no.~3, pp. 507--523, 2013.

\bibitem[Antico et~al.(2019)Antico, Sasazawa, Wu, Jaiprakash, Roberts,
  Crawford, Pandey, and Fontanarosa]{antico19us}
M.~Antico, F.~Sasazawa, L.~Wu, A.~Jaiprakash, J.~Roberts, R.~Crawford, A.~K.
  Pandey, and D.~Fontanarosa, ``Ultrasound guidance in minimally invasive
  robotic procedures,'' \emph{Medical Image Analysis}, vol.~54, pp. 149 -- 167,
  2019.

\bibitem[Long et~al.(2012)Long, Hungr, Baumann, Descotes, Bolla, Giraud,
  Rambeaud, and Troccaz]{long12prostate}
J.-A. Long, N.~Hungr, M.~Baumann, J.-L. Descotes, M.~Bolla, J.-Y. Giraud, J.-J.
  Rambeaud, and J.~Troccaz, ``{Development of a Novel Robot for Transperineal
  Needle Based Interventions: Focal Therapy, Brachytherapy and Prostate
  Biopsies.}'' \emph{{Journal of Urology}}, vol. 188, pp. 1369--1374, Aug.
  2012.

\bibitem[Boctor et~al.(2008)Boctor, Choti, Burdette, and Webster]{boctor08us}
E.~Boctor, M.~Choti, E.~Burdette, and R.~Webster, ``Three-dimensional
  ultrasound-guided robotic needle placement: An experimental evaluation,''
  \emph{International Journal of Medical Robotics and Computer Assisted
  Surgery}, vol.~4, no.~2, pp. 180--191, Jun. 2008.

\bibitem[Xu et~al.(2010)Xu, Jia, Song, Yang, Chen, and Liang]{xu10us}
J.~Xu, Z.-z. Jia, Z.-j. Song, X.-d. Yang, K.~Chen, and P.~Liang,
  ``Three-dimensional ultrasound image-guided robotic system for accurate
  microwave coagulation of malignant liver tumours,'' \emph{The International
  Journal of Medical Robotics and Computer Assisted Surgery}, vol.~6, no.~3,
  pp. 256--268, 2010.

\bibitem[Papalia et~al.(2012)Papalia, Simone, Ferriero, Costantini,
  Guaglianone, Forastiere, and Gallucci]{papalia12nephrectomy}
R.~Papalia, G.~Simone, M.~Ferriero, M.~Costantini, S.~Guaglianone,
  E.~Forastiere, and M.~Gallucci, ``Laparoscopic and robotic partial
  nephrectomy with controlled hypotensive anesthesia to avoid hilar clamping:
  Feasibility, safety and perioperative functional outcomes,'' \emph{Journal of
  Urology}, vol. 187, no.~4, pp. 1190--1194, 2012.

\bibitem[{Wu} et~al.(2015){Wu}, {Housden}, {Ma}, {Razavi}, {Rhode}, and
  {Rueckert}]{wu15catheter}
X.~{Wu}, J.~{Housden}, Y.~{Ma}, B.~{Razavi}, K.~{Rhode}, and D.~{Rueckert},
  ``Fast catheter segmentation from echocardiographic sequences based on
  segmentation from corresponding x-ray fluoroscopy for cardiac catheterization
  interventions,'' \emph{IEEE Transactions on Medical Imaging}, vol.~34, no.~4,
  pp. 861--876, 2015.

\bibitem[Issa et~al.(2019)Issa, Miller, and Zipes]{issa19mapping}
Z.~F. Issa, J.~M. Miller, and D.~P. Zipes, ``6 - advanced mapping and
  navigation modalities,'' in \emph{Clinical Arrhythmology and
  Electrophysiology (Third Edition)}, 3rd~ed., Z.~F. Issa, J.~M. Miller, and
  D.~P. Zipes, Eds.\hskip 1em plus 0.5em minus 0.4em\relax Philadelphia:
  Content Repository Only!, 2019, pp. 155 -- 205.

\bibitem[LaValle(2006)]{lavalle06planning}
S.~M. LaValle, \emph{Planning Algorithms}.\hskip 1em plus 0.5em minus
  0.4em\relax New York, NY, USA: Cambridge University Press, 2006.

\bibitem[Yushkevich et~al.(2006)Yushkevich, Piven, Cody~Hazlett, Gimpel~Smith,
  Ho, Gee, and Gerig]{py06nimg}
P.~A. Yushkevich, J.~Piven, H.~Cody~Hazlett, R.~Gimpel~Smith, S.~Ho, J.~C. Gee,
  and G.~Gerig, ``User-guided {3D} active contour segmentation of anatomical
  structures: Significantly improved efficiency and reliability,''
  \emph{Neuroimage}, vol.~31, no.~3, pp. 1116--1128, 2006.

\bibitem[Loschak et~al.(2016)Loschak, Degirmenci, Tenzer, Tschabrunn, Anter,
  and Howe]{loschak2016rob}
P.~M. Loschak, A.~Degirmenci, Y.~Tenzer, C.~M. Tschabrunn, E.~Anter, and R.~D.
  Howe, ``A four degree of freedom robot for positioning ultrasound imaging
  catheters,'' \emph{Journal of mechanisms and robotics}, vol.~8, no.~5, 2016.

\bibitem[Brattain et~al.(2014)Brattain, Loschak, Tschabrunn, Anter, and
  Howe]{brattain2014instrument}
L.~J. Brattain, P.~M. Loschak, C.~M. Tschabrunn, E.~Anter, and R.~D. Howe,
  ``Instrument tracking and visualization for ultrasound catheter guided
  procedures,'' in \emph{Workshop on Augmented Environments for
  Computer-Assisted Interventions}.\hskip 1em plus 0.5em minus 0.4em\relax
  Springer, 2014, pp. 41--50.

\bibitem[Enriquez et~al.(2018)Enriquez, Saenz, Rosso, Silvestry, Callans,
  Marchlinski, and Garcia]{enriquez2018use}
A.~Enriquez, L.~C. Saenz, R.~Rosso, F.~E. Silvestry, D.~Callans, F.~E.
  Marchlinski, and F.~Garcia, ``Use of intracardiac echocardiography in
  interventional cardiology: working with the anatomy rather than fighting
  it,'' \emph{Circulation}, vol. 137, no.~21, pp. 2278--2294, 2018.

\bibitem[Kim et~al.(2009)Kim, Hijazi, Lang, and Knight]{kim2009use}
S.~S. Kim, Z.~M. Hijazi, R.~M. Lang, and B.~P. Knight, ``The use of
  intracardiac echocardiography and other intracardiac imaging tools to guide
  noncoronary cardiac interventions,'' \emph{Journal of the American College of
  Cardiology}, vol.~53, no.~23, pp. 2117--2128, 2009.

\bibitem[Banchs et~al.(2010)Banchs, Patel, Naccarelli, and
  Gonzalez]{banchs2010intracardiac}
J.~E. Banchs, P.~Patel, G.~V. Naccarelli, and M.~D. Gonzalez, ``Intracardiac
  echocardiography in complex cardiac catheter ablation procedures,''
  \emph{Journal of interventional cardiac electrophysiology}, vol.~28, no.~3,
  pp. 167--184, 2010.

\bibitem[Bazaz and Schwartzman(2003)]{bazaz2003site}
R.~Bazaz and D.~Schwartzman, ``Site-selective atrial septal puncture,''
  \emph{Journal of cardiovascular electrophysiology}, vol.~14, no.~2, pp.
  196--199, 2003.

\bibitem[Bommer et~al.(1979)Bommer, Weinert, Neumann, Neef, Mason, and
  DeMaria]{bommer1979determination}
W.~Bommer, L.~Weinert, A.~Neumann, J.~Neef, D.~T. Mason, and A.~DeMaria,
  ``Determination of right atrial and right ventricular size by two-dimensional
  echocardiography.'' \emph{Circulation}, vol.~60, no.~1, pp. 91--100, 1979.

\bibitem[{Wang} et~al.(2020){Wang}, {Bie}, {Han}, and {Fang}]{wang20hysteresis}
X.~{Wang}, D.~{Bie}, J.~{Han}, and Y.~{Fang}, ``Active modeling and
  compensation for the hysteresis of a robotic flexible ureteroscopy,''
  \emph{IEEE Access}, vol.~8, pp. 100\,620--100\,630, 2020.

\end{thebibliography}
}

\end{document}